%% file: main.tex
\documentclass{article}

\usepackage[final,nonatbib]{neurips_data_2024}
\usepackage[square,sort,comma,numbers]{natbib}

\usepackage[utf8]{inputenc} 
\usepackage[T1]{fontenc}    
\usepackage{hyperref}       
\usepackage{url}            
\usepackage{booktabs}       
\usepackage{amsfonts}       
\usepackage{nicefrac}       
\usepackage{microtype}      
\usepackage{xcolor}         

\newcounter{subsubsubsection}[subsubsection]
\renewcommand{\thesubsubsubsection}{\thesubsubsection.\arabic{subsubsubsection}}
\newcommand{\subsubsubsection}[1]{%
  \par\medskip
  \refstepcounter{subsubsubsection}
  \noindent\textbf{\thesubsubsubsection\quad #1} 
  \par 
  \medskip
}

\input{notations}

\input{math_commands}

\title{Benchmarking PtO and PnO Methods in the Predictive Combinatorial Optimization Regime}

\author{%
  Haoyu Geng$^{1}$$^{\dag}$, Hang Ruan$^{1}$$^{\dag}$, Runzhong Wang$^{2}$, Yang Li$^{1}$, Yang Wang$^{3}$, Lei Chen$^{3}$, Junchi Yan$^{1}$\thanks{Correspondence author. $\dag$ denotes equal contribution. This work was in part supported by NSFC 92370201, 72342023.}\\
    $^{1}$Dept. of CSE \& School of AI \& Moe Key Lab of AI, Shanghai Jiao Tong University\\
    $^{2}$Massachusetts Institute of Technology\\
    $^{3}$Finvolution Group \\
\texttt{\{genghaoyu98, zzrh01, yanglily, yanjunchi\}@sjtu.edu.cn}\\
   \texttt{runzhong@mit.edu, \{wangyang09, chenlei04\}@xinye.com} \\
}

\begin{document}

\maketitle

\begin{abstract}
Predictive combinatorial optimization, where the parameters of combinatorial optimization (CO) are unknown at the decision-making time, is the precise modeling of many real-world applications, including energy cost-aware scheduling and budget allocation on advertising. Tackling such a problem usually involves a prediction model and a CO solver. These two modules are integrated into the predictive CO pipeline following two design principles: ``Predict-then-Optimize (PtO)'', which learns predictions by supervised training and subsequently solves CO using predicted coefficients, while the other, named ``Predict-and-Optimize (PnO)'', directly optimizes towards the ultimate decision quality and claims to yield better decisions than traditional PtO approaches. However, there lacks a systematic benchmark of both approaches, including the specific design choices at the module level, as well as an evaluation dataset that covers representative real-world scenarios. To this end, we develop a modular framework to benchmark 11 existing PtO/PnO methods on 8 problems, including a new industrial dataset for combinatorial advertising that will be released. Our study shows that PnO approaches are better than PtO on 7 out of 8 benchmarks, but there is no silver bullet found for the specific design choices of PnO. A comprehensive categorization of current approaches and integration of typical scenarios are provided under a unified benchmark. Therefore, this paper could serve as a comprehensive benchmark for future PnO approach development and also offer fast prototyping for application-focused development. The code is available at \url{https://github.com/Thinklab-SJTU/PredictiveCO-Benchmark}.

\end{abstract}

\section{Introduction}\label{sec:intro}

\begin{figure*}[tb!]
    \centering
    \includegraphics[width=\textwidth]{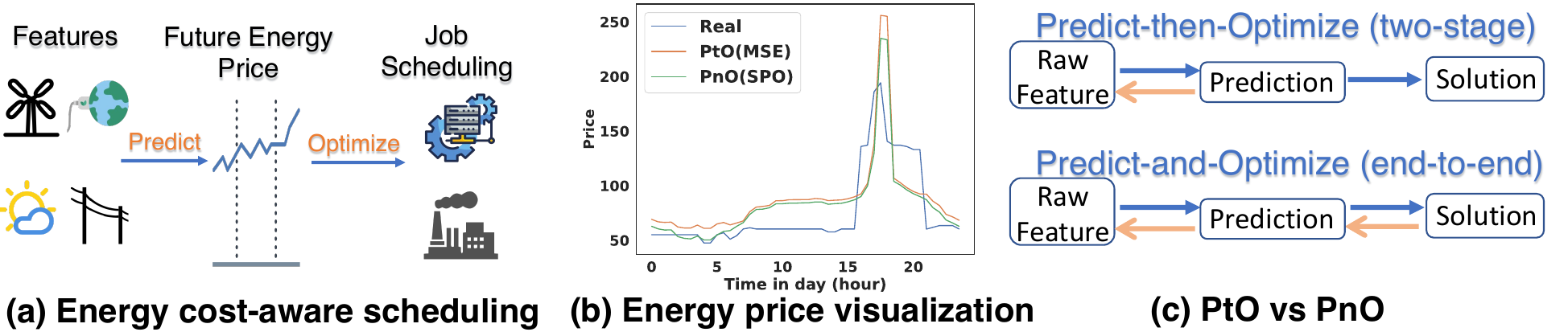}
    \vspace{-10pt}
    \caption{  (a) Example of predictive CO in energy-cost aware scheduling. Factories deploy job schedules based on energy price predictions to reduce production costs. (b) Visualization of energy prices of the 200th test instance in SEMO dataset. PtO makes improper predictions, which further prescribes sub-optimal decisions. (c) PtO vs. PnO. PnO designs decision-oriented training approaches that emerged recently as a promising direction to tackle predictive CO.}
    \label{fig:intro}
    \vspace{-10pt}
\end{figure*}

Predictive combinatorial optimization is a family of Combinatorial Optimization (CO) problems where the problem parameters are unknown during decision-making. Predictive CO models a wide range of real-world settings, including energy cost-aware scheduling~\cite{wahdany2023more},  budget allocation for website information dissemination~\cite{wilder2019melding}, portfolio optimization~\cite{markowitz2000mean}. In predictive CO, the optimization itself could be solved by readily available solvers (e.g. Gurobi~\cite{gurobi2019llc}) once the problem parameters are predicted. However, due to the inherent uncertainty and noise from the real world, the prediction is never perfect, resulting in potentially misleading problem parameters. Even the optimal solution under these parameters could be a bad decision in the real world. As shown in Fig~\ref{fig:intro}(a\textasciitilde b), job scheduling without considering energy price fluctuations may result in energy waste and higher costs.\looseness=-1

\looseness=-1 Hence, there are important and practical needs to jointly consider both prediction and decision-making for predictive optimization (especially CO) problems~\cite{elmachtoub2022smart,mandi2020smart,cameron2022perils}. 
A basic solution for predictive CO is ``\textbf{predict-then-optimize}''~\cite{bertsimas2020predictive} (\textbf{PtO}, or called the two-stage approach), which first predicts coefficients of the optimization task through a predictive model trained under the supervision of ground truth coefficients and subsequently utilizes off-the-shelf solvers to obtain solutions. Intuitively, it is expected that a higher prediction accuracy would result in better decision quality. Nonetheless, as also evidenced by Figure~\ref{fig:finetune}(c\textasciitilde d), there often exists a gap between prediction objectives and ultimate decision goals, leading to suboptimal decisions by PtO models. Therefore, as shown in Fig~\ref{fig:intro}(b), a recent series of studies~\cite{mandi2020smart,wang2020automatically} have proposed the new ``\textbf{predict-and-optimize}'' paradigm (\textbf{PnO}, or decision-focused learning~\cite{wilder2019melding,mandi2022decision,shah2022decisionfocused} that learns to directly cater for the ultimate decision objectives. In cases where the quality of the final decision is paramount, joint prediction-and-optimization approaches may be more direct and beneficial. Though this can sometimes come at the expense of prediction accuracy (as is shown in Fig~\ref{fig:finetune}(c)), they exhibit notable improvement over PtO on several optimization tasks. 



The main challenges towards predict-and-optimize for predictive CO lie in two aspects: (1) the solution's derivatives with respect to the optimization coefficients are not available. (2) the variables in the optimization problems are discrete. Both challenges lead to the blocking of gradients.
To mitigate this issue, several approaches have been proposed, including designing proxy loss functions~\cite{mandi2020smart}, gradient approximation~\cite{poganvcic2019differentiation}, etc. However, there lacks a systematic categorization of existing methods, and it is not clear which methods are effective for specific problems and scenarios. Moreover, the experimental benchmarks are relatively fragmented, and the existing proposed models have not undergone comprehensive evaluation, impeding the community's progression. Though some implementations~\cite{agrawal2019differentiable,tang2022pyepo} are available for some methods, they only support linear or convex problems, while overlooking other non-linear and submodular problems. Moreover, the optimization datasets are limited in scale, and some of them use generated data for the prediction part, lacking an industrially large-scale real-world dataset for validating the PnO performance. 

 In this work, we systematically review existing methodologies and establish a benchmark to align each model. Based on empirical results, we then conduct a series of in-depth experiments to explore key factors for PnO model design, and find several useful tips for deployment of PnO models.
Based on these findings, we provide a set of recommendations for practitioners on model selection when considering the applicability and practical performance.
We anticipate that our open-source benchmark and new dataset will gain increased attention, thereby fostering advancements in both research and practical applications.
Our contributions and conclusions are summarized as follows:

\bb$\quad$We systematically review current PnO approaches and categorize them into four categories according to how the problem is solved regarding the decision variables: discrete or continuous; and how the loss function is designed: (statistically) direct or surrogate. We also give a PnO model choice recommendation by analyzing the potential gains and challenges during the real-world deployment.

\bb$\quad$ We develop a modular framework of 11 existing PtO/PnO methods on 8 problems, and multiple solvers for a fair benchmarking. We also include a new industrial dataset regarding the combinatorial advertising problem, formulated as an integer linear program with uncertain conversion rates. 


\bb$\quad$ Our benchmark results demonstrate PnO is better than PtO on 7 out of 8 predictive CO problems. However, \textbf{no silver bullet is found for specific design choices}, suggesting the necessity of trial-and-error for various scenarios. Therefore, our comprehensive and modular-based benchmark covering mainstream PnOs could further help quick prototyping in application-focused development.

\bb$\quad$ Our extensive benchmark discovers some key factors for PnO methodology design, suggesting future research directions. Specifically, we try to answer 3 research questions regarding \textbf{relationship between prediction accuracy and decision quality}, \textbf{impacts of prediction labels on PnO}, and \textbf{versatility of PnO across different settings}. The experiments indicate that leveraging decision information for more favorable trade-offs may be a key factor in how PnO works; pre-training with predicted labels can enhance certain PnO methods; the versatility of PnO in optimization parameters still needs improvement.


\begin{table*}[tb!]
\caption{Categorization of representative methods, where ``Predictive'' belongs to ``PtO'' while others are ``PnO''s. BB, ID, CPLayer is short for Blackbox, Identity, and cvxpy layer. {``\halfcheckmark'' means it supports the characteristic in some models among many. Suppose $N$ is the decision variable size, and $K$ is the number of extra optimization samples. $T_{LODL}$ denotes backward time following ~\cite{shah2022decisionfocused}.}
}
\label{tab:cmp}
\scalebox{0.61}{
\begin{tabular}{l c c c ccc c ccc c c c cc}
\toprule
& {\textbf{Predictive}} && \multicolumn{3}{c}{\textbf{Discrete}}      && \multicolumn{3}{c}{\textbf{Continuous}} &&{\textbf{Statistical}}  &&  \multicolumn{2}{c}{\textbf{Surrogate}}  \\ 

\cline{2-2} \cline{4-6} \cline{8-10} \cline{12-12} \cline{14-15}

On-the-fly         & {\Checkmark} && \multicolumn{3}{c}{\Checkmark}    && \multicolumn{3}{c}{\Checkmark} && \multicolumn{1}{c}{\XSolidBrush} && {\XSolidBrush} & \halfcheckmark \\

$\partial \mathbf{z}/\partial \mathbf{y}$          & None   && Innate       & \multicolumn{2}{c}{Interpolation} && \multicolumn{2}{c}{Automatic differentiation}     & Designed       && \multicolumn{1}{c}{Statistical}  && \multicolumn{2}{c}{Learned}    \\

Approach & \textbf{Two-stage}       && \textbf{DFL}
&\textbf{BB}
& \textbf{ID}
&&\textbf{QPTL}
& \textbf{CPLayer}
& \textbf{SPO}
&&\textbf{NCE}, 
\textbf{LTR}
&& \textbf{LODL}
& \textbf{SurCO}
\\
Cite & \cite{bertsimas2020predictive} && \cite{shah2022decisionfocused}    
& \cite{poganvcic2019differentiation}  
& \cite{sahoo2022backpropagation}   
&&\cite{wilder2019melding} 
& \cite{agrawal2019differentiable} 
& \cite{mandi2020smart}    
&&\cite{ijcai2021p0390},\cite{mandi2022decision} 
&& \cite{shah2022decisionfocused} 
& \cite{ferber2023surco} \\

Objective           & Any    && Any     & Linear           & Linear         && Quadratic & Convex    & Linear && \multicolumn{1}{c}{Any}          && {Any}   & {Nonlinear}     \\

Constraints & Any  && Any  & Linear  & Linear   && Linear & Convex  & Linear && {Any}  &&  {Any}  & {Integer}   \\

{Requires $y^{\star}$} &  \Checkmark  && \Checkmark    &  \XSolidBrush  & \XSolidBrush     &&  \XSolidBrush   &\XSolidBrush & \Checkmark  && \Checkmark && \Checkmark & \halfcheckmark \\

Additional solving & 0   && 0    & 1   & 0     && 0    & 0   & 1 && {$K$}     && {$K$}  & {$K\cdot |\mathcal{D}|$}  \\ 

{\small{Backward Complexity}} &  $O(1)$  && $O(1)$  &  $O(N)$  & $O(1)$  &&  $O(N^{3})$ & $O(N^{3})$  & $O(N)$  && $O(KN)$ && $O(K\cdot T_{LODL})$  & $O(N^2)$ \\ 
\bottomrule
\end{tabular}%
}
\end{table*}

\section{Problem Formulation}\label{sec:problem}
 Consider a predictive combinatorial optimization with unknown coefficients $\rvy$:
\begin{equation}
\max_{\mathbf{z} \in \mathcal{Z}}\quad f(\mathbf{z}, \mathbf{y}, \bm{\theta})~ s.t.~ \mathbf{z} \in \operatorname{Constr}(\bm{\theta}) ~,
\end{equation}
 where $f(\mathbf{z},\mathbf{y},\bm{\theta})$ denotes the known and closed-formed optimization objective (abbreviated as $f(\mathbf{z},\mathbf{y})$ below) with discrete variable $\mathbf{z} \in \mathcal{Z}$, and $\mathbf{y}$ is collection of unknown optimization coefficients, $\bm{\theta}$ are optimization parameters that are known and fixed, and decision variables $\mathbf{z}$ obey the constraints $\operatorname{Constr}(\bm{\theta})$. We assume that the coefficients of the constraints are known in line with the majority of literature~\cite{wilder2019melding,shah2022decisionfocused,mandi2022decision}. We denote the solution obtained through one solver call for a problem instance with coefficient $\mathbf{y}$ as $\mathbf{z}(\mathbf{y})$, and $\mathbf{z}^{\star}(\mathbf{y})$ as the optimal solution. The solvers can be commercial solvers (e.g. Gurobi~\cite{gurobi2019llc}), open-sourced solvers (e.g. cvxpy~\cite{diamond2016cvxpy}), or neural solvers (e.g. submodular~\cite{karimi2017stochastic}).

Though coefficients $\mathbf{y}$ are unknown, in many circumstances, they could be estimated by a predictive model using a collection of historical or pre-collected dataset $\mathcal{D}=\{(\mathbf{x}_i, \mathbf{y}_i)\}$, where $\mathbf{x}$ denotes relevant raw features. Therefore, the learning objective of the prediction step is:
\begin{equation}
\min_{\gM} \mathbb{E}_{(\mathbf{x}_i, \mathbf{y}_i) \sim \mathcal{D}}[\mathcal{L}_{pred}(\hat{\mathbf{y}}_i, \mathbf{y}_i)] ~,
\end{equation}
where $\mathcal{L}_{pred}$ is a training loss specified by the prediction output, e.g. mean squared error (MSE). Suppose the prediction model $\mathcal{M}$ serves as a mapping from the feature vector $\mathbf{x}_i$ to coefficients $\hat{\mathbf{y}}_i$, i.e. $\hat{\mathbf{y}}_i = \mathcal{M}(\mathbf{x}_i)$, the predictive optimization problem~\cite{wilder2019melding} is:
\begin{equation}
\underset{\mathbf{z}_i\in \mathcal{Z}}{\max} \quad \mathbb{E}_{(\mathbf{x}_i, \mathbf{y}_i) \sim \mathcal{D}}\left[f\left(\mathbf{z}_i(\mathcal{M}(\mathbf{x}_i)), \mathcal{M}(\mathbf{x}_i), \bm{\theta}\right)\right].
\end{equation}

The evaluation of the PtO could be critical. In many circumstances where the real coefficients of the test set are available, regret~\citep{mandi2020smart,yan2021Surrogate,guler2022divide,mandi2022decision} is used to evaluate decision quality. Let the decision quality~\cite{shah2022decisionfocused,ferber2023surco} of solution $\hat{\mathbf{z}}$ be the objective under the ground-truth coefficient $\mathbf{y}$:
\begin{equation}\label{eq:dq}
\operatorname{DQ}(\hat{\mathbf{z}}) = f(\hat{\mathbf{z}}, \mathbf{y},\bm{\theta}).
\end{equation}

Then the regret could be obtained by the difference of the decision quality with solutions under the estimated coefficient($\mathbf{z}^{\star}(\mathbf{y}),$) and ground-truth coefficients ($\mathbf{z}^{\star}(\hat{\mathbf{y}})$):
\begin{equation}
    \operatorname{Regret}(\hat{\mathbf{y}}, \mathbf{y}) = |f(\mathbf{z}^{\star}(\mathbf{y}),\mathbf{y},\bm{\theta}) - f(\mathbf{z}^{\star}(\hat{\mathbf{y}}),\mathbf{y},\bm{\theta})|.
\end{equation}
However, in many cases (e.g. combinatorial advertising problem), the ground truth coefficients $\rvy$ are not readily available or even impossible to obtain, so the evaluation of PtO becomes challenging. Online A/B testing may be one approach; however, it is associated with higher costs and risks in practice. Consequently, we propose using uplift, an offline controlled variable metric common in causal inference tasks, as an alternative evaluation elaborated in Appendix~\ref{suppl:uplift}.

\section{Categorization of Existing Methods}\label{sec:method}
\begin{wrapfigure}[25]{R}{0.5\textwidth}
\vspace{-15pt}
\centering
\includegraphics[width=.45\columnwidth]{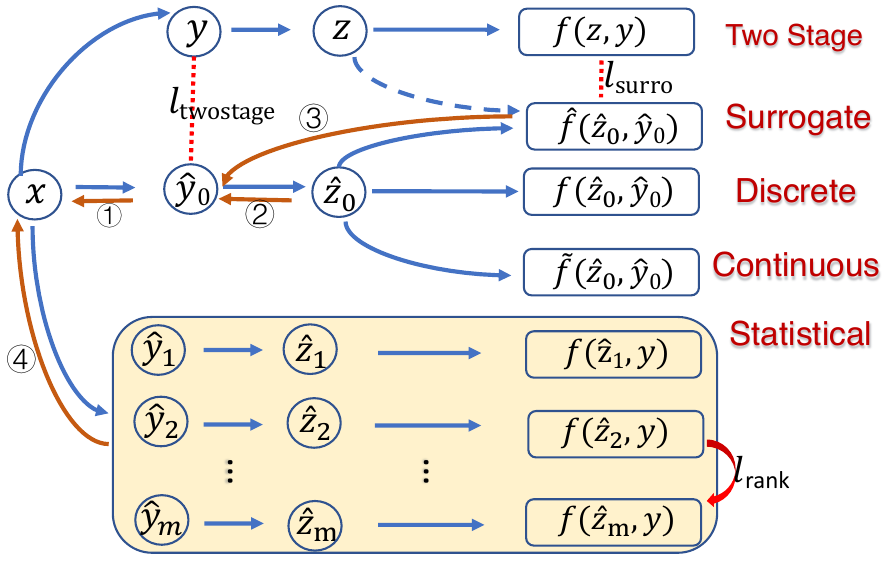}
\caption{Gradient flow of existing PnO methods, where $f$, $\hat{f}$ and $\tilde{f}$ denote the original objective, learned objective by surrogate model, and the continuous relaxation of the original objective, respectively. The path of back propagation of the vanilla two-stage is \ding{172}, while discrete/continuous categories use~\ding{172}\ding{173}, statistical one uses~ \ding{175}, and surrogate one uses~ \ding{172}\ding{174}. $l_{surro}$ represents the surrogate losses to measure how the surrogate imitates the original optimization objective, while $l_{rank}$ refers to the designed loss that encodes solution ranking—e.g., ensuring that the optimal solution is assigned a lower loss than suboptimal ones.} 
\label{fig:grad}
\end{wrapfigure}

 As briefly introduced earlier, the major challenges of PnO lie in two aspects: (1) unavailable/non-informative gradients and (2) discrete variables. These two challenges both block the gradient backpropagation so that the prediction model cannot be updated by the final objective. Table~\ref{tab:cmp} categorizes existing works into four groups by how the optimization is solved and how the gradient is obtained, where the gradient flow of these approaches is shown in Figure~\ref{fig:grad}. 
We consider the \textbf{PtO} training as a baseline method while concurrently noting that the first challenge can be readily addressed if the solver is differentiable, such as neural solvers for top-k~\cite{xie2020differentiable} or Sinkhorn algorithm~\cite{sinkhorn1964relationship} used in neural graph matching~\cite{WangICCV19}. 
The technical details are given in Appendix~\ref{suppl:pno-details}.

\begin{figure*}[tb!] 
\centering 
\includegraphics[width=0.95\textwidth]{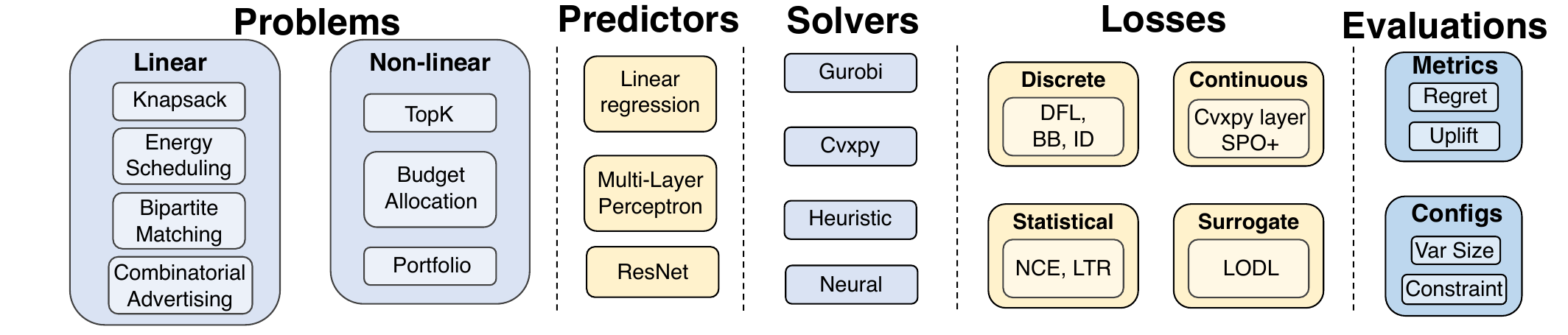}
\caption{A modular code framework supporting 11 problems, 8 PtO/PnO models, multiple solvers, and various evaluations under configurable parameters. Users can easily customize their own problems, predictors, models, and solvers.} 
\label{fig:code}
\vspace{-10pt}
\end{figure*}

\subsection{The Discrete Category}

 As discussed in~\cite{poganvcic2019differentiation,berthet2020learning,niepert2021implicit}, the gradient of decision variable $\mathbf{z}$ with respect to coefficients $\mathbf{y}$ revealed as piecewise constant functions (almost zero everywhere, and undefined otherwise).
To address this issue, the ``discrete'' category solves the optimization with the original discrete solver in the forward pass, while in the backward pass, the gradients are estimated via designed interpolation functions, such as linear interpolation done by Blackbox~\cite{poganvcic2019differentiation}. 
Representative models include \textbf{DFL}~\cite{shah2022decisionfocused}, \textbf{Blackbox}~\cite{poganvcic2019differentiation}, \textbf{Perturb}~\cite{berthet2020learning} and \textbf{I-MLE}~\cite{niepert2021implicit}, \textbf{Identity}~\cite{sahoo2022backpropagation}, etc.

\looseness=-1 One advantage of this category is that the gradient interpolations work on the fly and do not require coefficient labels or additional optimization samples. Therefore, this class of methods requires the least additional resources.
However, they are only adaptable to linear or convex optimization objectives, since gradients of more complex objectives are hard to estimate.
Though a vanilla DFL method is agnostic to the optimization form, it is prone to have high variances~\cite{shah2022decisionfocused} and our experiments also indicate such issues (shown in Figure~\ref{fig:sensi-cap}). Besides, some methods require additional solver calls in the forward pass, like BlackBox, Perturb, and I-MLE, which adds a bit of computational overhead.

\subsection{The Continuous Category}

 The ``continuous'' category estimates the gradient of the relaxed version of the CO problem since continuous problems are implicitly differentiable in nature. 
In the forward pass, it generally solves the relaxed optimization problem, whereas in the backward pass, the gradients are obtained by automatic differentiation (of such as KKT conditions). 
Representative models include \textbf{OptNet}~\cite{amos2017optnet},  \textbf{CPLayer}~\cite{agrawal2019differentiable} and more~\cite{donti2017task,lee2019meta,eisenberger2022unified} 
for continuous PnOs and 
\textbf{QPTL}~\cite{wilder2019melding},
\textbf{IntOpt}~\cite{mandi2020interior}, SPO-relax~\cite{mandi2020smart} (short as SPO below), and more~\cite{wang2020automatically,ferber2020mipaal,paulus2021comboptnet}, for PnO on discrete problems.

 The advantage of this category is that the gradient $\partial z/\partial y$ is readily available when the problem is relaxed to continuous. However, it necessitates tailored relaxations for each individual problem, and not every discrete problem has readily available or straightforward relaxations. Additionally, the application of KKT (or subgradient)-based differentiation is limited to convex (or linear) objectives, rendering it not suitable for all CO problems.

\subsection{The Statistical Category}

 The ``statistical'' category~\cite{ijcai2021p0390,mandi2022decision} designs loss for PnO considering the statistical relation between multiple solutions. The aim is to make coefficient predictions so that the decision quality (defined in Eq.~(\ref{eq:dq})) of optimal solutions is superior to suboptimal solutions.
Representative models include \textbf{NCE}~\cite{ijcai2021p0390} and \textbf{LTR}~\cite{mandi2022decision} (with 3 versions: pointwise-LTR, pairwise-LTR, and listwise-LTR).

\looseness=-1 The statistical category is usually agnostic to the problem type and solver. However, a solution cache of optimization samples is required before the training, which creates additional overhead. It is also not readily applicable to problem instances of varying variable sizes for solution cache in implementation.

\subsection{The Surrogate Category}

 Another branch besides ``statistical'' that bypasses gradient estimation is the ``surrogate'' category. It generally replaces the original optimization objective with a learned surrogate function and has emerged as effective in recent literature. It is inherently differentiable since the objective function is learned.
Representative works include \textbf{LODL}~\cite{shah2022decisionfocused}, \textbf{EGL}~\cite{shah2024leaving},
\textbf{LANCER}~\cite{zharmagambetov2023landscape}, 
\textbf{SurCO}~\cite{ferber2023surco}, etc.



 Similar to the statistical category, the surrogate category easily adapts to all problem and solver types. Furthermore, in cases where solving is slow or assessing solution quality is expensive, using the surrogate expedites the PnO training process. Nevertheless, they do necessitate an extra set of optimization instances and corresponding solutions to learn objective functions as initialization. Additionally, whether and how well they can learn complex nonlinear functions remains open.

\subsection{Takeaways for Practitioners' PnO Model Choice}\label{sec:guide}

\begin{wrapfigure}[14]{L}{0.48\textwidth}
\vspace{-12pt}
\centering
\includegraphics[width=.43\columnwidth]{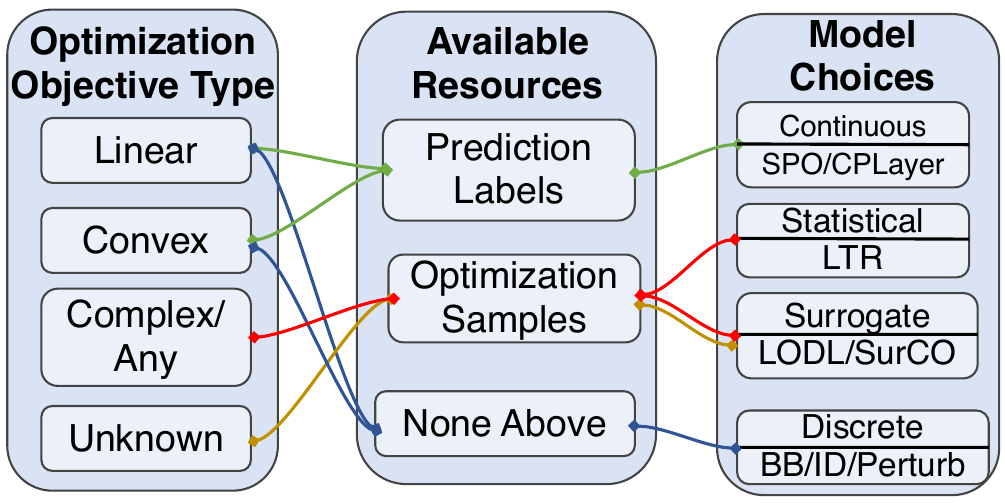}
\caption{Choice guide of PnO models, which depends on optimization objective type, available resources. From top to bottom, it becomes harder to deploy PnO for harder optimization types and fewer available resources.} 
\label{fig:guide}
\end{wrapfigure}

\textbf{Remark on 4 categories of PnO methods}: 
As categorized in Table~\ref{tab:cmp}, while end-to-end training on PnO has emerged as a novel paradigm with promising potential to surpass the two-stage PtO paradigm,  in practice, there does not currently exist a singular end-to-end model capable of outperforming all other methods: the discrete-category and continuous-category are efficient with fewer solver calls, while constrained by certain optimization forms (e.g. convex), while the latter two categories fit any optimization objectives but with high computational cost and requires additional optimization samples.

\looseness=-1 \textbf{PnO model choice guideline} As shown in Fig~\ref{fig:guide}, we consider several key factors for the preferable model recommendation: requirement for \textit{labeled prediction datasets}, \textit{extra optimization samples}, and \textit{overall performance} (in both efficacy and efficiency).
\textbf{Firstly}, the SPO~\cite{mandi2020smart} method is preferable for most linear or near-linear problems (Knapsack, energy-cost aware scheduling, budget allocation problem, etc.) as the first try, as long as there are labeled datasets for prediction tasks. SPO requires fewer labeled datasets and is generally faster regarding additional solving, backward pass complexity, etc.
\textbf{Secondly}, the LTR~\cite{mandi2022decision} method is a good choice for non-linear tasks (TopK, bipartite matching, Portfolio optimization) as long as there are extra optimization samples at hand (or corresponding solution cache), and LODL~\cite{shah2022decisionfocused} for better performance if optimization solutions are easy to get (since it requires abundant samples and solutions).
\textbf{Finally}, the discrete category (Blackbox~\cite{poganvcic2019differentiation}/Identity~\cite{sahoo2022backpropagation}) is the last choice for linear/convex problems if there are no labeled datasets for prediction tasks, such as combinatorial advertising. 

\section{Benchmarks and Remarks}\label{sec:experi}
\subsection{Modular Framework Design}
As shown in Figure~\ref{fig:code}, we implement the benchmark as a modular framework where each scenario is composed of problems, predictors, solvers, losses, and evaluations. This enables a comprehensive and convenient evaluation of current methods. 
Although we have not exhaustively addressed all CO problems, we have selected representative ones from real-world applications and introduced a real-world industrial dataset to contribute to the field's development.

\begin{table*}[tb!]
\caption{Dataset statistics. Short terms: integer linear program (ILP), quadratic program (QP). 
}
\label{tab:data-all}
\scalebox{.75}{%
\begin{tabular}{lccccccc}
\toprule
Name               &  \makecell[c]{Data\\Source} & \makecell[c]{Variable\\Size}  &  \makecell[c]{Train\\Samples} &  \makecell[c]{Test\\Samples}&  \makecell[c]{Optimization\\Type} &  \makecell[c]{Used\\Solver} & \makecell[c]{Prediction\\Loss}\\  \midrule

Energy scheduling & SEMO~\citep{ifrim2012properties} & 48                 &    650        &     139  & ILP       & Gurobi~\citep{gurobi2019llc} &  MSE  \\
Knapsack (Gen) & Generated~\cite{elmachtoub2022smart} & 20  & 400  &   200 & ILP       & Gurobi~\citep{gurobi2019llc} &   MSE    \\
Knapsack (Energy) & SEMO~\citep{ifrim2012properties}  & 48   &  400     &    200    & ILP      & Gurobi~\citep{gurobi2019llc}  &  MSE\\
Cubic Top-k       & Generated~\citep{shah2022decisionfocused} & 50  & 250 & 400  & Top-K  & Heuristic~\cite{paszke2019pytorch}  &  MSE \\
Budget allocation  & Yahoo~\citep{yahoo07} & 100   & 400 &  200 & Submodular & Submodular~\citep{karimi2017stochastic}    & BCE \\
Bipartite matching &  Cora~\citep{sen2008collective}  & 50 &  20  & 6 & ILP & CVXPY~\citep{diamond2016cvxpy}   &  BCE  \\
Portfolio optimization & Quandl~\citep{quandl} & 50  & 400 & 200 & QP & CVXPY~\citep{diamond2016cvxpy} &  MSE\\
 \makecell[l]{Combinatorial Advertising} & \makecell[c]{{Industrial}} &  \makecell[c]{2933(avg)} & 23 & 6 & ILP  &  Ortools~\citep{ortools}  &  BCE \\       
\bottomrule
\end{tabular}
}
\end{table*}

\subsection{Problem Descriptions}\label{sec:dataset}
In this section, we briefly introduce each task with the datasets in Table~\ref{tab:data-all}, where the background, practical scenarios, detailed ``prediction'' and ``optimization'' formulations are left in Appendix~\ref{suppl:data-detail}.

\subsubsection{Benchmark on publicly available datasets}
\vspace{-5pt}
\textbf{Energy-cost Aware Scheduling} (\textbf{\textit{SE}})
The energy cost-aware scheduling task adopted from~\cite{mandi2020smart,mandi2020interior,ijcai2021p0390,guler2022divide,mandi2022decision}, is to make machine production schedules based on the predicted energy prices.

\textbf{Knapsack} (\textbf{\textit{KG}}, \textbf{\textit{KE}})
We consider a knapsack problem with unknown item values used in~\cite{demirovic2019investigation,mandi2020smart,mandi2020interior,ijcai2021p0390,guler2022divide}.
We provide two versions: the synthetic version~\cite{elmachtoub2022smart,tang2022pyepo}(knapsack (gen), or ``\textbf{\textit{KG}}''), and a real energy dataset version~\cite{mandi2020interior,ijcai2021p0390,guler2022divide} (knapsack (energy), or ``\textbf{\textit{KE}}'').

\textbf{Budget Allocation} (\textbf{\textit{BA}})
\textbf{\textit{BA}} has applications when agents intend to disseminate information~\cite{wilder2019melding,shah2022decisionfocused} across multiple websites to reach a wider audience within a budget constraint, but the probability of each website reaching users is unknown.

\textbf{Cubic Top-K} (\textbf{\textit{TK}})
The Top-k finds $K$ largest numbers with unknown values following~\cite{shah2022decisionfocused}.

\textbf{Bipartite Matching} (\textbf{\textit{BM}})
We consider a bipartite graph matching used in ~\cite{wilder2019melding,ferber2020mipaal,mandi2022decision} where the edge connectivity is unknown and requires prediction.

\textbf{Portfolio Optimization} (\textbf{\textit{PF}})
We introduce portfolio optimization~\cite{donti2017task,wang2020automatically} with unknown asset prices, which maximizes the immediate net profit of the securities while reducing risk. Though a continuous quadratic program, we bring it here since it is often regarded as a stress test~\cite{shah2022decisionfocused} where the quadratic program naturally provides informative gradients by automatic differentiation~\cite{amos2017optnet}.

\subsubsection{New dataset of combinatorial advertising for inclusive finance}
\vspace{-5pt}
 
Extending financial services to specific underserved populations within society is highly beneficial in mitigating wealth disparities and enhancing the living standards of low-income individuals. 
We introduce a new dataset of Combinatorial Advertising (``\textbf{\textit{CA}}'') of real industry advertising records, in which a fintech platform connects with financial institutions to provide low-interest loans to users. 

 \textbf{Dataset}: The advertisement takes place on a mobile application (APP) as a combination of various channels, including in-app notifications, text messages, telephones, etc. Historical data contains whether a user converted after being exposed to a specific marketing combination in the past; however, labels for other combinations remain unknown. 
The data can be accessed through the website\footnote{\url{https://opendata.xinye.com}}, and the processing and use terms are in Appendix~\ref{suppl:data-adv-detail}.

\textbf{Prediction}: 
Given user $i$'s feature $\mathbf{x}^i$ (encrypted and processed personal features, app activity records, etc.), predict the user's conversion rate $\mathbf{y}^{ij}$ to the $j$-th combination of advertising channels.

\textbf{Optimization}:
The goal is to allocate the advertiser's existing budget to offer each user a combination that enables broader access (higher conversions) of users to financial services.

 It differs from the above \textbf{\textit{BA}} problem as \textbf{\textit{CA}} is based on personalized advertising while \textbf{\textit{BA}} is not. 
\textbf{\textit{CA}} is also more challenging than the multiple treatment setting~\cite{zhou2023direct} due to the exponential combination space of channels, where the latter refers to multiple levels of treatments of the same channel.



\subsection{Experimental Setup}\label{sec:setup}
We implement our framework based on the previous works~\cite{mandi2022decision,shah2022decisionfocused,tang2022pyepo}. All experiments are carried out on a workstation with Intel$^{\circledR}$ i9-7920X, NVIDIA$^{\circledR}$ RTX 2080, and 128GB RAM.  

For the predictive model, we use a two-layer MLP with 32 hidden units in each layer. 
Unless otherwise specified, we extract 20\% from the training data set for validation.
During training, we adopt the Adam Optimizer~\cite{kingma2015adam} and search by grid the learning rate in $\{0.1,,0.05,0.01,0.005,0.001\}$. 
In each run, each model is trained for 300 epochs and the training stops if no better regret is achieved for 50 epochs on the validation set. The epoch that reaches the lowest regret (or highest uplift) in validation is selected for testing. We elaborate on more details of the model design in Appendix~\ref{suppl:model-detail}.

\begin{table*}[tb!]
\caption{Results for 7 problems on 11 methods. The relative regret (w.r.t. optimal objective \%) and average runtime of each epoch are reported. ``-'' means non-applicable and data generation time of LODL is not counted. Best three: \red{red}, \orange{orange}, \blue{blue}.}
\label{tab:bench}
\scalebox{.58}{%
\begin{tabular}{clccccccccccccccc}
\toprule
     Problem &  & {\textbf{Predictive}}  & & \multicolumn{3}{c}{\textbf{Discrete}}   & & \multicolumn{2}{c}{\textbf{Continuous}} &  & \multicolumn{4}{c}{\textbf{Statstical}} & & {\textbf{Surrogate}}  \\ 
      \cline{3-3} \cline{5-7} \cline{9-10} \cline{12-15} \cline{17-17}
      
      &   & {Two-stage} && {DFL} & {Blackbox} & {Identity} && {CPLayer}     & {SPO} && {NCE} & {point-LTR} & {pair-LTR} & {list-LTR} && {LODL} \\ \midrule
      
\multirow{3}{*}{\makecell[c]{Knapsack\\(Gen)}}  
        & Regret (\%)    & 6.595 & & 11.744 &  24.274 & 31.874 & & 24.769 & \blue{6.223} & & 13.438 & 6.402 & 7.820 & \red{6.031} & & \orange{6.044} \\
        & Train Time & 0.101 && 1.208 &  2.195 & 1.351 & & 0.368 & 1.173 & & 2.065 & 1.846 & 4.861 & 1.536 & & {0.344} \\
        & Test Time      & 0.794 && 1.636 &  1.544 & 1.553 & & 1.928 & 1.100 & & 1.445 & 1.682 & 3.973 & 1.119 & & {0.716} \\
        \hline

\multirow{3}{*}{\makecell[c]{Knapsack\\(Energy)}}
        & Regret (\%)     & 8.745 & & \blue{8.353} & 35.705 & 17.156 & & 36.402 & 8.407 & & 11.932 & \orange{8.236} & 9.022 & \red{8.083} & & {9.567}\\ 
        & Train Time & 0.198 & &1.116 & 1.776 & 1.051 & & 0.416 & 2.007 & & 4.010 & 2.602 & 10.396 & 3.099 & & {0.501}\\
        & Test Time      & 0.342 && 0.760  & 0.752 & 0.745 & & 1.825 & 1.069 & & 2.054 & 1.406 & 6.751 & 1.699 & & {0.344} \\
        \hline

\multirow{3}{*}{\makecell[c]{Scheduling\\(Energy)}} 
        & Regret (\%)&1.793 && 6.272  & 6.503 & 5.690 & & - & \red{1.505} & & 1.663 & 4.548 & \orange{1.540} & \blue{1.551} & & {1.786} \\
        & Train Time & 0.404 && 51.857 & 110.647 & 56.334 & & - & 110.362 & & 109.645 & 65.744 & 61.048 & 61.873 & & {0.582} \\
        & Test Time  & 13.196 && 26.186 &  26.799 & 26.920 & & - & 41.448 & & 37.904 & 28.590 & 27.100 & 26.945 & & {12.055} \\
        \hline 
                          
\multirow{3}{*}{\makecell[c]{Budget\\Allocation}} 
        & Regret (\%) & 20.332 && 35.970 & 26.905 & 14.799 & & - & \red{5.559} & & 9.979 & 69.663 & \blue{5.958} & \orange{5.742} & & {25.700} \\
        & Train Time & 0.102 && 4.019 & 39.799 & 21.224 & & - & 40.021 & & 7.327 & 22.669 & 23.194 & 22.774 & & {0.248}  \\
        & Test Time  & 12.828 && 25.138  & 25.728 & 26.344 & & - & 38.127 & & 38.460 & 26.392 & 26.293 & 26.168 & & {12.278} \\
        \hline 

\multirow{3}{*}{\makecell[c]{TopK\\(Cubic)}} 
        & Regret (\%)  & \red{0.110} && 1.974 & 13.944 & 13.944 & & - & 160.408 & & 160.408 & 1.149 & 5.072 & \blue{0.193} & & \orange{0.172} \\
        & Train Time & 0.064 && 0.116 & 0.096 & 0.097 & & - & 0.126 & & 0.393 & 0.379 & 4.653 & 0.679 & & {0.197} \\
        & Test Time   & 0.038 && 0.105 & 0.090 & 0.087 & & - & 0.125 & & 0.890 & 0.832 & 11.625 & 1.571 && {0.034} \\
        \hline 

\multirow{3}{*}{\makecell[c]{Bipartite\\Matching}} 
        & Regret (\%) & 92.963 && \blue{91.364} & 91.988 & 91.868 & & 92.007 & 93.327 & & 92.622 & \red{91.035} & 92.285 & 91.831 & & \orange{91.113}\\
        & Train Time   & 0.010 & & 25.121 & 0.623 & 0.334 & & 17.179 & 1.895 & & 3.467 & 0.659 & 1.501 & 1.531 & & {0.051} \\
        & Test Time   & 0.262 & & 7.725 & 0.446 & 0.457 & & 6.643 & 1.160 & & 1.758 & 0.886 & 1.055 & 1.044 & & {0.239} \\
        \hline
                          
\multirow{3}{*}{Portfolio} 
        & Regret    & \blue{0.243} & & 0.380 & 0.286 & 0.280 & & 0.309 & 0.245 & & 0.367 & \orange{0.214} & 0.255 & 0.249 & & \red{0.160} \\
        & Train Time & 0.204 & & 0.754 & 3.762 & 2.018 & & 1.121 & 3.446 & & 11.082 & 3.382 & 16.383 & 3.564 & & {0.198} \\
        & Test Time      & 1.187 && 3.547 & 2.251 & 3.115 & & 3.504 & 3.121 & & 8.071 & 3.829 & 17.087 & 3.919 & & {1.182} \\         
\bottomrule                 
\end{tabular}
}
\end{table*}

\subsection{Benchmark Results}\label{sec:benchmark}
\vspace{-5pt}

We list the benchmarks, including relative regret (with respect to optimal objective) and average runtime for training and test in Table~\ref{tab:bench}, and hyper-parameter sensitivity in Appendix~\ref{suppl:exp-sensi}.

From the perspective of each method, methods in the discrete category do not perform as well as others, especially on non-linear optimization tasks (budget allocation, topk, portfolio, etc.). This is probably because the gradient interpolations are only designed for linear objectives and gradient estimations in the back-propagation may not be accurate in complex problems. 
The continuous-category method SPO performs well on linear problems with relatively short training time. We run CPLayer on the linear optimization problems and non-linear programs are omitted. 
For the statistical category, listwise-LTR achieves better results than others in most problems, probably because it captures the global relationships among multiple solutions.
The surrogate method LODL achieves satisfactory results on half of the benchmarks.
More analysis is left to Appendix~\ref{suppl:bench-res}.

As for the training efficiency, one may observe that the two-stage approach is significantly faster than others in the training stage since it does not involve solving the optimization problem. CPLayer may take much longer time than others since the KKT differentiation may incur $O(N^3)$ time complexity for back-propagation. Statistical methods often come with higher training time since multiple solutions are required for model initialization.  Though training on LODL is fast, the model preparation time (including collecting solutions for optimization samples and learning the surrogate models for objective functions) is not counted, and it could be time-consuming if the number of required samples is large and solving an optimization problem is slow. We used 5000 samples for LODL, and better LODL performance may occur with more samples (as well as more running time)

\begin{wraptable}[7]{L}{0.43\textwidth}
\vspace{-15pt}
\caption{Combinatorial advertising of ``discrete'' category.}
\label{tab:bench-adv}
\scalebox{0.7}{%
\begin{tabular}{lcccc}
\toprule
  &{Two-stage} & {DFL} & {Blackbox} & {Identity} \\ \midrule
     Uplift     & 0.069 & 0.088 & 0.134 & 0.135  \\ 
     Train Time & 0.170 & 10.670 & 10.053 &  10.120  \\
     Test Time  &  3.078 &  3.216 &  3.096 &  3.199  \\
  \bottomrule 
\end{tabular}
}
\end{wraptable}

The result of the combinatorial advertising on discrete category is shown in Table~\ref{tab:bench-adv}, where other categories of methods that cannot run on-the-fly do not apply to this problem directly because they require additional coefficient labels/optimization samples that cannot be satisfied in this problem. We conclude applicability in Sec.~\ref{sec:guide} in the following. 
The PnO training significantly improves the uplift metric (using the discrete category). The identity model achieves the best result.
The DFL and Blackbox approach also performs better than the two-stage approach, though with higher training time.
Though the end-to-end PnO training takes higher training time, they do not incur additional runtime during the test stage.

\textbf{Remark of PnO methods by empirical results}: In the benchmark results, we observe that in line with analysis in Sec.~\ref{sec:method}, despite the immediate applicability of discrete-category methods without extra data or solving, their reliance on interpolated gradients often leads to inaccuracies during training, resulting in inferior performance compared to the two-stage methods in many scenarios. Statistical-category and surrogate-category methods exhibit better performance. However, it is noteworthy that they incur higher computational costs during data sample collection and forward-pass computations. 

The above categorizations and benchmark suggest that \textit{PnO methods lack a universally accepted standard, and in practical applications, each method has its own advantages and disadvantages}. Therefore, there is an urgent need for better approaches with respect to applicability, efficiency and performance. Hence, addressing the imperative challenge of attaining a more \textbf{universal}, \textbf{stable}, and \textbf{efficient} training model is essential to fully realize the potential of ``PnO'' in the real world.

\begin{figure*}[tb!]
    \centering
\begin{adjustbox}{width=\textwidth}
    \begin{tabular}{cccc}
        \hspace{-1cm} 
        \includegraphics[width=0.25\columnwidth]{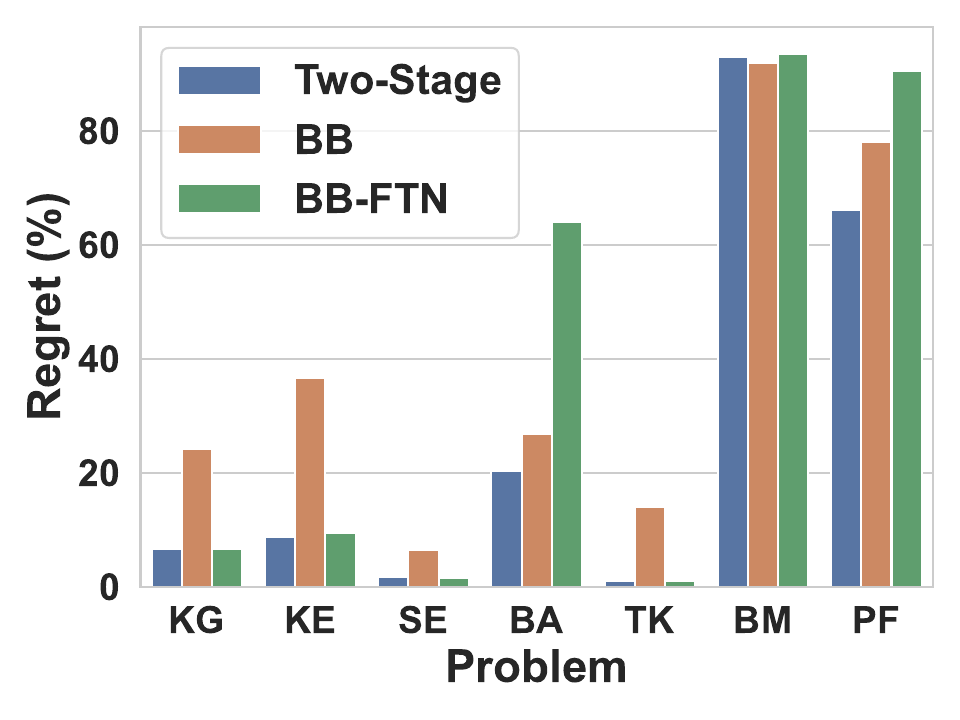}
        &\includegraphics[width=0.25\columnwidth]{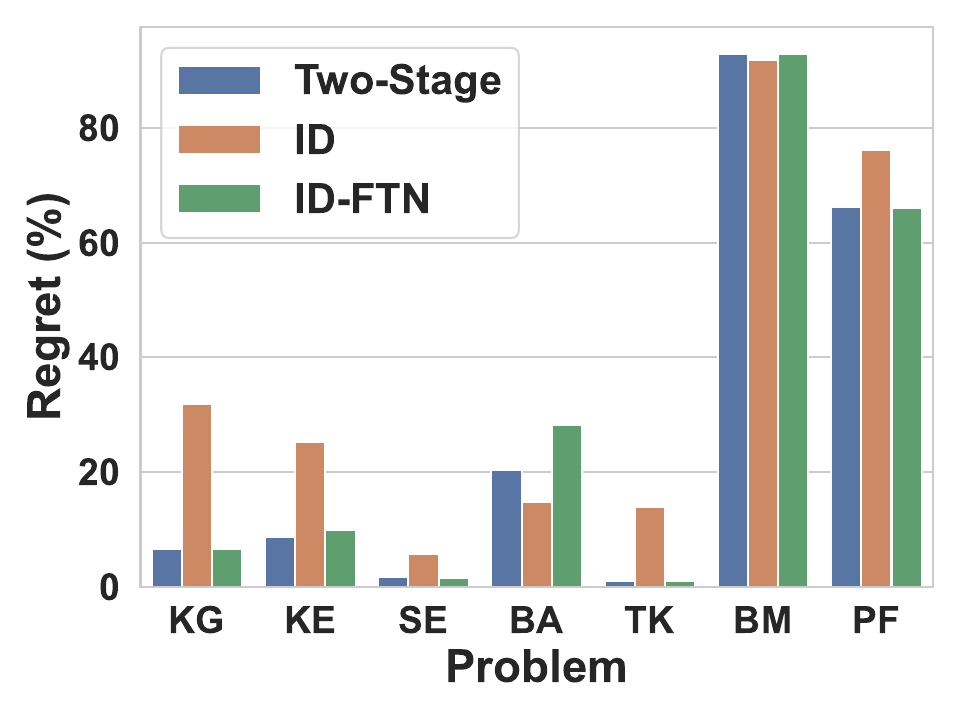} 
        &\includegraphics[width=0.25\columnwidth]{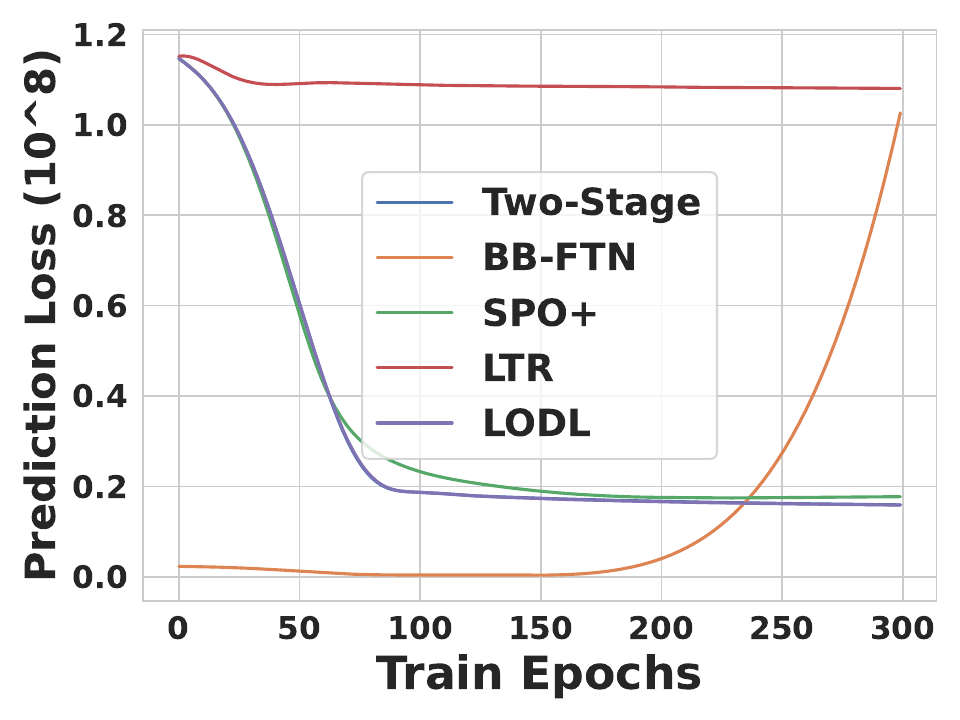}
        &\includegraphics[width=0.25\columnwidth]{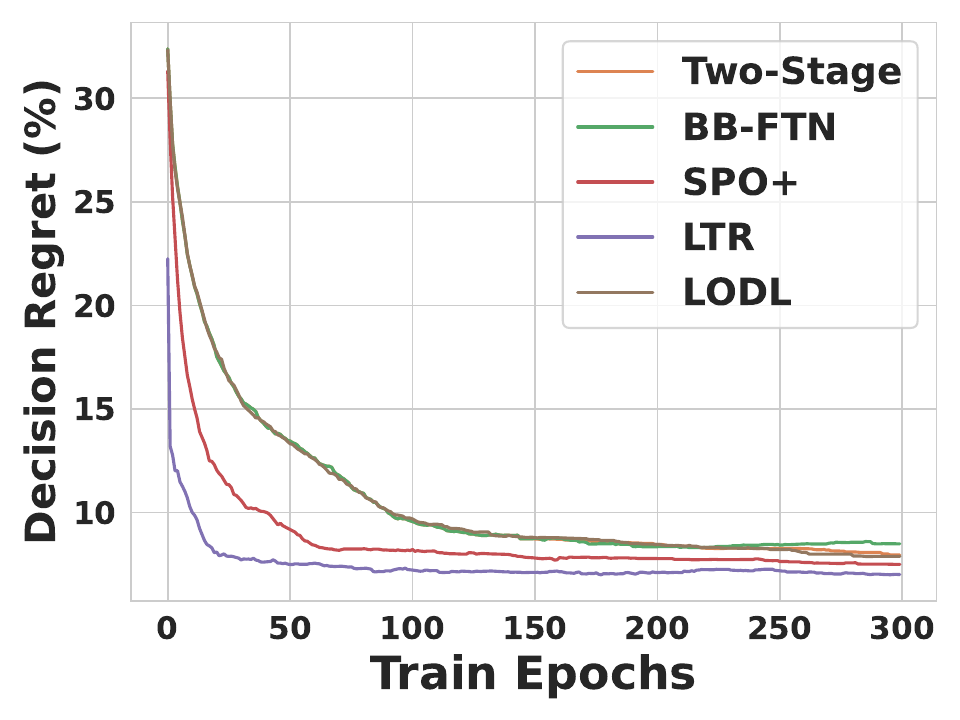} 
        \\  
    (a) ``fine-tuned'' BB model & (b) ``fine-tuned'' ID model & (c) Prediction loss on training  & (d) Regret on training
    \end{tabular}
 \end{adjustbox}
 \vspace{-5pt}
    \caption{(a\textasciitilde b) Results of fine-tuning (short as ``FTN'') of the discrete category (BB, ID model). Compared to training PnO directly, ``BB-FTN'' and ``ID-FTN'' benefit by first pertaining by PtO and then fine-tuning by PnO on 4,5 over 7 datasets, respectively.  (c\textasciitilde d) Learning curve {on knapsack (gen) dataset} for prediction loss and decision regret w.r.t. training epochs. PnO approaches (LTR, SPO) achieve lower regret than the PtO approach (Two-stage), though with higher prediction loss.}
    \label{fig:finetune}
\end{figure*}


\subsection{Key Factors for PnO Methodology Design}
To investigate the key factors for future improvements of PnO, we propose to explore the following research questions: (\textbf{RQ1}) What is the relationship between prediction accuracy and decision quality in the PnO methods? (\textbf{RQ2}) To what extent do the prediction labels impact the PnO methods? (\textbf{RQ3}) Do the PnO methods demonstrate versatility across different settings?

\textbf{RQ1: Relationships of prediction accuracy and decision quality} In common sense, a better prediction would lead to a better subsequent decision. Besides a case study shown in \cite{elmachtoub2022smart}, we explore the prediction loss and decision objective in both PtO and PnO. We illustrate the learning curve on the real-world knapsack (energy) dataset by plotting the prediction loss (Mean-squared error) and evaluation results (regret) of one method for each category, shown in Figure~\ref{fig:finetune}(c~d). 

 It is observed that PnO training methods like SPO, listwise-LTR, and LODL achieve lower regret than the vanilla PtO method. The prediction loss of LODL shares a similar pattern with PtO, while LTR's is different. This demonstrates despite higher prediction loss incurred in the end-to-end training, these PnO methods obtain better decisions by their own loss functions concerning the ultimate decision objectives.
This might correspond to the proposition~\cite{cameron2022perils} that \textit{PnO achieves better decision quality by finding better error trade-offs by information of decision objectives}. We show more qualitative analysis in Appendix~\ref{suppl:quali}.

\textbf{RQ2: Impact of prediction labels on PnO} 
We explore the impacts of prediction labels specifically for discrete categories (Blackbox and Identity) since they inherently do not use label information. 
Without the true prediction label $\mathbf{Y}$, training from scratch by respective PnO loss could make it very slow to converge, as a possible reason for demonstrating inferior performance in Table~\ref{tab:bench}.
We attempt to expose the model with true prediction labels as a ``warm-up'' before the PnO training: we initially train for 150 epochs using a PtO MSE loss, followed by 150 epochs of fine-tuning using the PnO loss. The results are denoted as ``BB-FTN'' and ``ID-FTN'' respectively, where ``FTN'' is short for fine-tuning. 
It could be observed from Figure~\ref{fig:finetune}(a~b) that this hybrid training on Blackbox and Identity method demonstrates improvements over trained directly by PnO across 4 and 5 over 7 datasets, respectively, and surpasses the two-stage method in some cases. 
This may indicate that \textit{supervised PtO pretraining could be beneficial for PnO models, especially for the discrete category that inherently does not rely on prediction labels} (see Table~\ref{tab:cmp} for requirements for prediction labels).

\begin{figure*}[tb!]
    \centering
\begin{adjustbox}{width=\textwidth}
    \begin{tabular}{ccc}
        \includegraphics[width=0.25\textwidth]{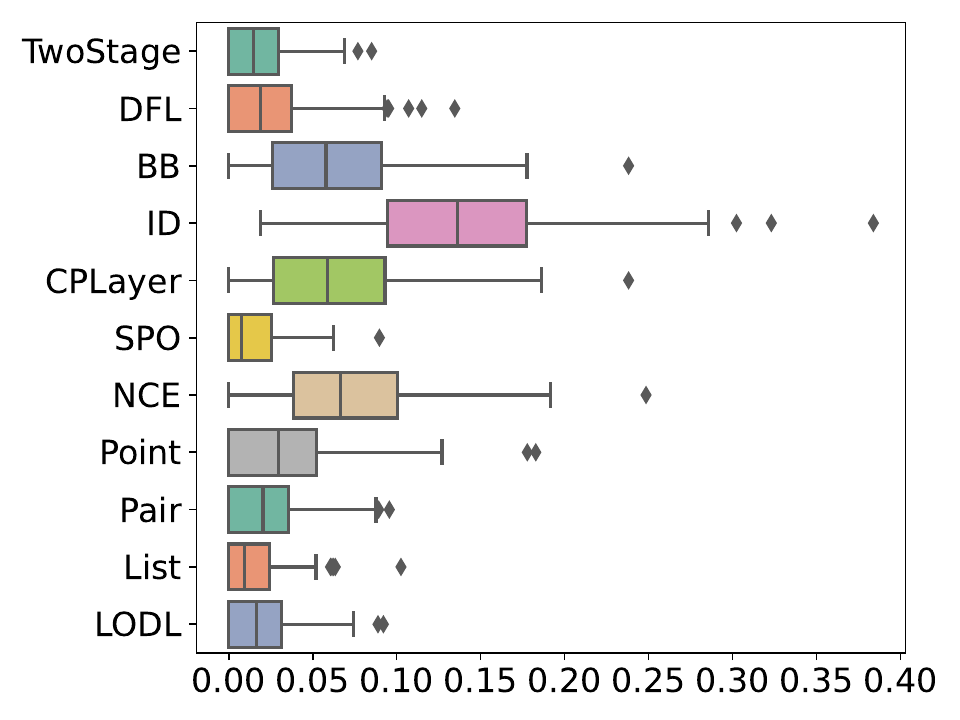}
        &\includegraphics[width=0.25\textwidth]{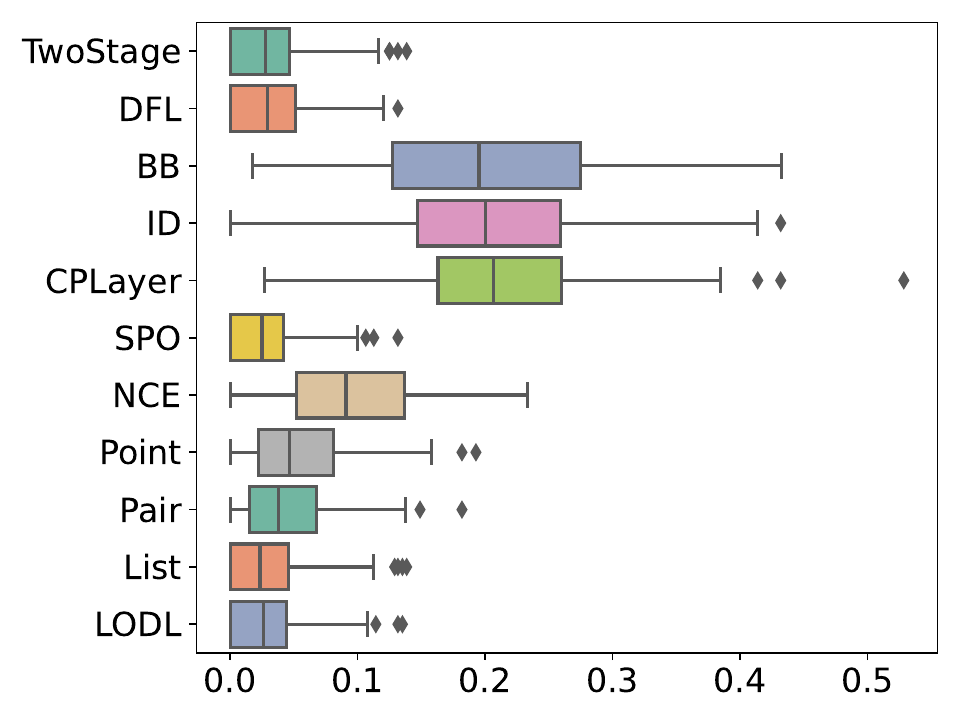} 
        &\includegraphics[width=0.25\textwidth]{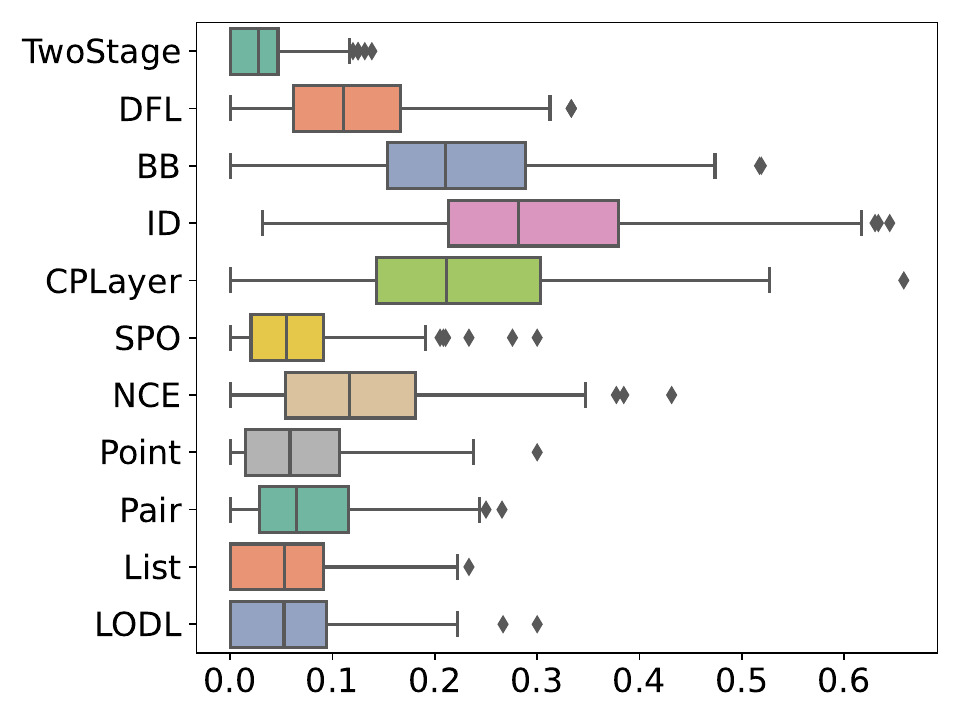}     
        \\  
    (a) Variable size 20, Capacity 90 & (b)  Variable size 20, Capacity 60 & (c)  Variable size 20, Capacity 30
    \end{tabular}
\end{adjustbox}
    \vspace{-10pt}
    \caption{Regret on knapsack (gen) problem. The decision degrades with tighter constraints.}
    \label{fig:sensi-cap}
\end{figure*}

\begin{figure*}[tb!]
    \vspace{-10pt}
    \centering
    \begin{adjustbox}{width=\textwidth}
    \begin{tabular}{cccc}
        \hspace{-0.7cm} 
        \includegraphics[width=0.25\textwidth]{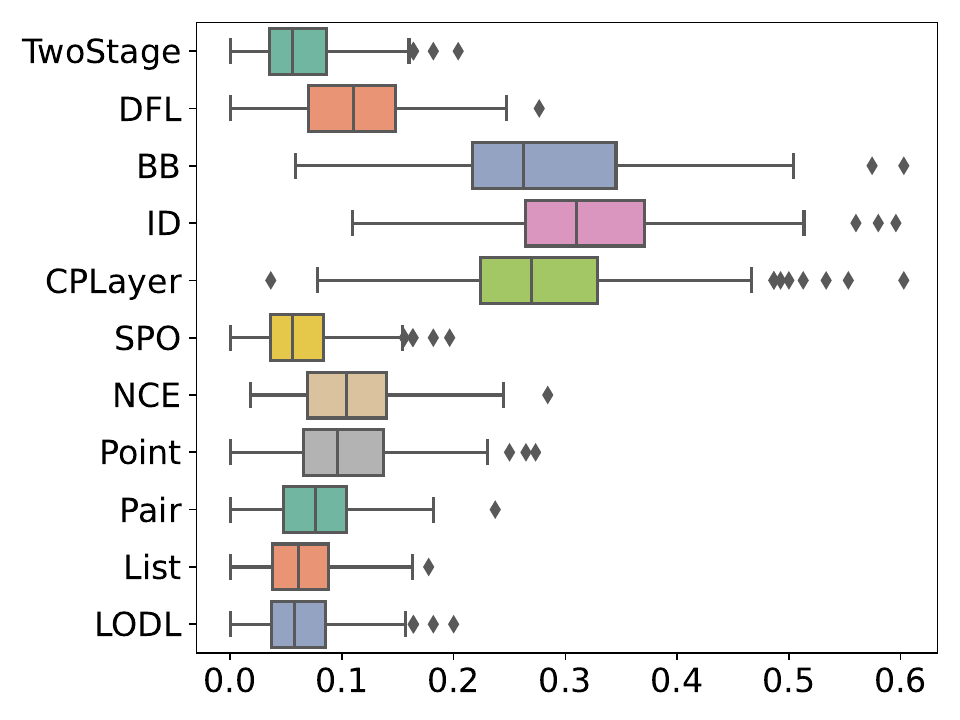}
        &\includegraphics[width=0.25\textwidth]{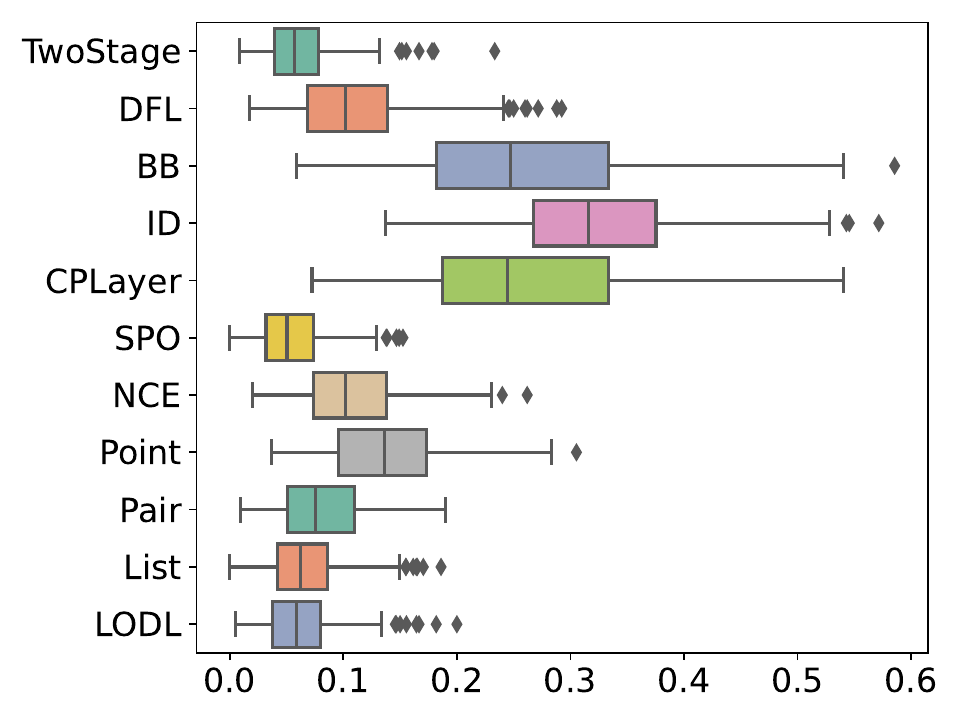} 
        &\includegraphics[width=0.25\textwidth]{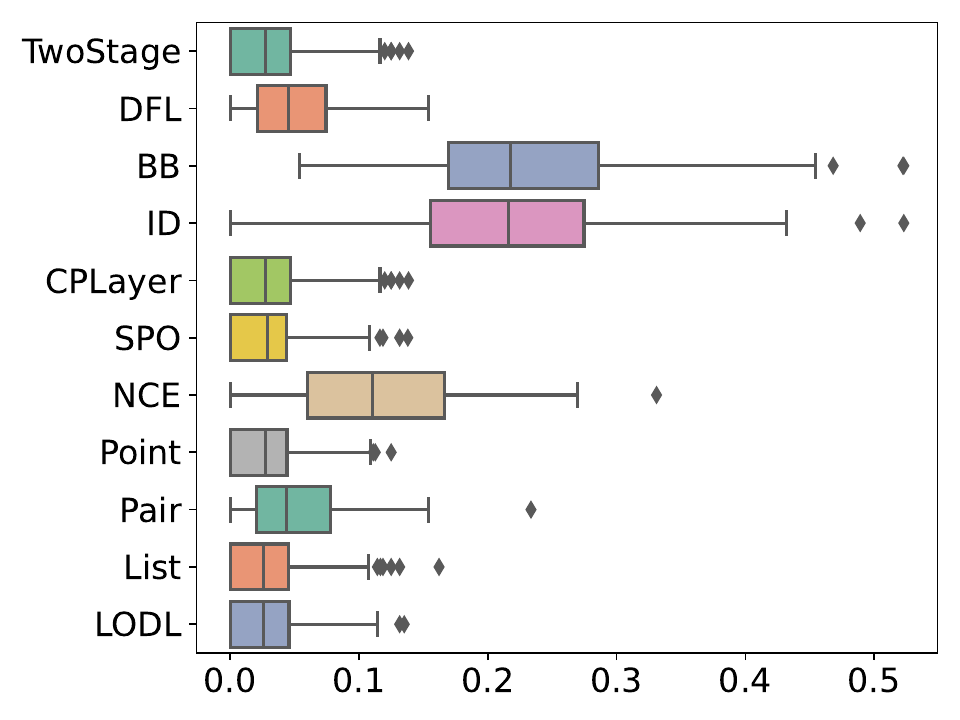} 
        &\includegraphics[width=0.25\textwidth]{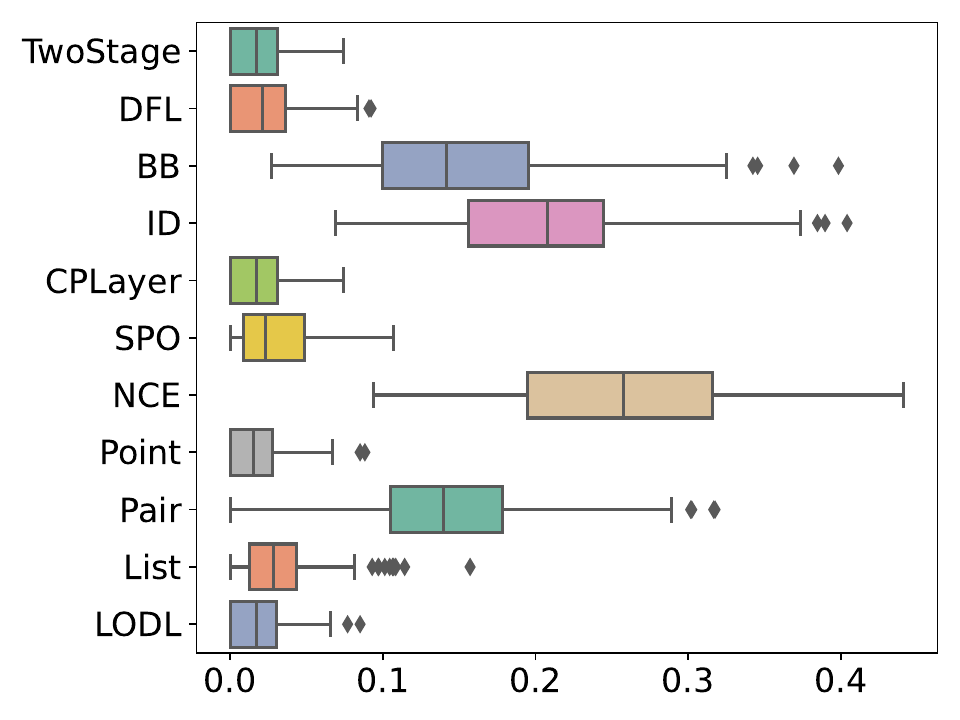} 
        \\  
    (V1) Variable size 40 & (V2) Variable size 60 & (G1) Generalize to cap 60 & (G2) Generalize to cap 90  
    \end{tabular}
\end{adjustbox}
    \vspace{-10pt}
    \caption{Regret on knapsack (gen) problem. (V1)\textasciitilde (V2) demonstrates performance with different variable sizes, and (G1)\textasciitilde (G2) generalizes the trained model on capacity 30 to 60, 90 on the test data. Most PnO models degrade when the test set distribution shifts and are less stable than the two-stage approach with different variable sizes. Detailed results in Table.~\ref{tab:sensitivity}.}
    \label{fig:sensi-size_generalize}
    \vspace{-10pt}
\end{figure*}

\textbf{RQ3: Versatility of PnO under various parameters} 
We explore the versatility of PnO as follows.
 (\textbf{1}) Figure~\ref{fig:sensi-cap} examines the impact of varying constraint on the training of \textbf{\textit{KG}} dataset. We set the capacity of 30, 60, and 90. It was observed that as the constraints became more restrictive, the relative regrets became higher. This observation potentially indicates that constraint satisfaction is a prevalent factor constraining the performance of current PnO training methods.
(\textbf{2}) Figure~\ref{fig:sensi-size_generalize}(V1\textasciitilde V2) show the performance when the number of decision variables increases to 40 and 60. With increasing variable size, the two-stage approach remains stable under different problem settings, though its final decision quality is inferior to some of the PnO training models. Conversely, some PnO models exhibited fluctuations and degradation, such as discrete-category methods and point-wise LTR. (\textbf{3}) Figure~\ref{fig:sensi-size_generalize}(G1\textasciitilde G2) show the generalizability when the PnO models are trained on one problem setting (e.g., constraint of capacity 30) and tested on the other (e.g., constraint of capacity 60 or 90). Similar performance degradation occurs in some PnO models, including discrete-category and statistical-category models, in this setting. This indicates \textit{current PnO approaches may perform inadequately when facing more stringent constraints, larger decision variable sizes, and the need for generalization across diverse optimization parameters}.
More details are shown in Appendix~\ref{supp:fails}.

\section{Conclusion}\label{sec:conclusion}
 Regarding predictive CO problems, we systematically review existing PtO/PnO methods, implement a framework for comprehensive assessment, and perform experiments to explore scenarios in which PnO methods could surpass the PtO approaches. Furthermore, we introduce a novel dataset, which will be publicly available along with the framework. We hope to facilitate the community and industry in swiftly reproducing, developing, and deploying new algorithms in the real world. 

\bibliographystyle{abbrv}
\bibliography{ref_opt,ref_graph}

\clearpage
\appendix
\input{appendix}

\clearpage
\section*{Checklist}


\begin{enumerate}

\item For all authors...
\begin{enumerate}
  \item Do the main claims made in the abstract and introduction accurately reflect the paper's contributions and scope?
    \answerYes{The main claims in the abstract and introduction accurately reflects the paper's contributions and scope.}
  \item Did you describe the limitations of your work?
    \answerYes{We discuss the limitations in Appendix~\ref{suppl:limits}.}
  \item Did you discuss any potential negative societal impacts of your work?
    \answerYes{We discuss the impacts in Appendix~\ref{suppl:limits}.}
  \item Have you read the ethics review guidelines and ensured that your paper conforms to them?
    \answerYes{We read the ethics review guidelines and ensure the paper conforms to them.}
\end{enumerate}

\item If you are including theoretical results...
\begin{enumerate}
  \item Did you state the full set of assumptions of all theoretical results?
    \answerNA{We make some attempts to explain the benchmark results with existing theories in Appendix~\ref{suppl:theory}, but it is not our primary focus.}
	\item Did you include complete proofs of all theoretical results?
    \answerNA{We make some attempts to explain the benchmark results with existing theories in Appendix~\ref{suppl:theory}, but it is not our primary focus.}
\end{enumerate}

\item If you ran experiments (e.g. for benchmarks)...
\begin{enumerate}
  \item Did you include the code, data, and instructions needed to reproduce the main experimental results (either in the supplemental material or as a URL)?
    \answerYes{We specified the model details in Sec.~\ref{sec:setup} and Appendix~\ref{suppl:model-detail}. The code will be open-sourced after publication.}
  \item Did you specify all the training details (e.g., data splits, hyperparameters, how they were chosen)?
    \answerYes{We specify training details in Appendix~\ref{suppl:data-detail} and \ref{suppl:exp-detiail}}
	\item Did you report error bars (e.g., with respect to the random seed after running experiments multiple times)?
    \answerYes{We report error bars for knapsack problem in Fig~\ref{fig:sensi-size_generalize}.}
	\item Did you include the total amount of compute and the type of resources used (e.g., type of GPUs, internal cluster, or cloud provider)?
    \answerYes{We specify the used computing resource in Sec~\ref{sec:setup}.}
\end{enumerate}

\item If you are using existing assets (e.g., code, data, models) or curating/releasing new assets...
\begin{enumerate}
  \item If your work uses existing assets, did you cite the creators?
    \answerYes{We cite the dataset sources in Table~\ref{tab:data-all}, and specified all the dataset details in Appendix~\ref{suppl:data-detail}.}
  \item Did you mention the license of the assets?
    \answerYes{We discuss the license of assets in Appendix~\ref{suppl:data-detail}.}
  \item Did you include any new assets either in the supplemental material or as a URL?
    \answerYes{We introduce our new dataset in Appendix~\ref{suppl:data-adv-detail}.}
  \item Did you discuss whether and how consent was obtained from people whose data you're using/curating?
    \answerYes{We discuss in Appendix~\ref{suppl:data-adv-lic}.}
  \item Did you discuss whether the data you are using/curating contains personally identifiable information or offensive content?
    \answerYes{We discuss in Appendix~\ref{suppl:data-adv-lic}.}
\end{enumerate}

\item If you used crowdsourcing or conducted research with human subjects...
\begin{enumerate}
  \item Did you include the full text of instructions given to participants and screenshots, if applicable?
    \answerNA{The paper does not use crowdsourcing or conducted research with human subjects.}
  \item Did you describe any potential participant risks, with links to Institutional Review Board (IRB) approvals, if applicable?
    \answerNA{The paper does not use crowdsourcing or conducted research with human subjects.}
  \item Did you include the estimated hourly wage paid to participants and the total amount spent on participant compensation?
    \answerNA{The paper does not use crowdsourcing or conducted research with human subjects.}
\end{enumerate}

\end{enumerate}


\end{document}

%% file: notations.tex
\usepackage[utf8]{inputenc} 
\usepackage{booktabs}       
\usepackage{amsfonts}       
\usepackage{nicefrac}       
\usepackage{microtype}      
\usepackage{xcolor}         

\usepackage{microtype}
\usepackage{graphicx}
\usepackage{subfigure}
\usepackage{booktabs} 
\usepackage{amsmath}
  
\usepackage{amssymb}
\usepackage{mathtools}
\usepackage{amsthm}
\usepackage{enumitem, multirow, xspace}
\usepackage{amsmath, amsfonts}
\usepackage{csquotes}
\usepackage{siunitx}
\usepackage{float}
\usepackage{bbm}
\usepackage{wrapfig}
\usepackage{lipsum}
\usepackage{bm}
\usepackage{arydshln}

\usepackage{makecell}
\usepackage{balance}
\usepackage{threeparttable}
\newcommand{\nop}[1]{}
\newcommand{\bb}{\hspace{-1mm} $\bullet$}
\usepackage{dsfont}
\usepackage{bbding}
\usepackage{pifont}
\usepackage{wasysym}
\usepackage{adjustbox}
\theoremstyle{plain}

\theoremstyle{definition}

\theoremstyle{remark}

\usepackage{algpseudocode}  
\usepackage[linesnumbered,ruled,vlined]{algorithm2e}
\SetKwInOut{Parameter}{Parameters}

\usepackage{tikz}

\def\halfcheckmark{\textcolor{black}{\ding{51}}{\small\textcolor{black}{\kern-0.7em\ding{55}}}}

\newcommand{\red}[1]{{\color{red} \textbf{#1}}}

\newcommand{\orange}[1]{{\color{orange}{#1}}}
\newcommand{\blue}[1]{{\color{cyan} \textbf{#1}}}

%% file: math_commands.tex
\usepackage{amsmath,amsfonts,bm}

\def\eqref#1{equation~\ref{#1}}

\def\1{\bm{1}}

\def\rvy{{\mathbf{y}}}

\DeclareMathAlphabet{\mathsfit}{\encodingdefault}{\sfdefault}{m}{sl}
\SetMathAlphabet{\mathsfit}{bold}{\encodingdefault}{\sfdefault}{bx}{n}

\def\gM{{\mathcal{M}}}



%% file: appendix.tex
\appendix
\section{Related Work}\label{sec:related}

Despite the significant progress neural networks have made in machine learning, there is still substantial room for improvement when it comes to solving optimization problems. Although several studies~\cite{qiu2022dimes,jin2023pointerformer,sun2023difusco,chen2024efficient,hertrich2023provably,li2023t2t,li2024fastt2t,li2023hardsatgen,chen2024mixsat,guo2024acm} have explored the use of neural networks for combinatorial optimization problems in deterministic environments, especially in the constrained CO tasks~\cite{wang2023linsatnet,wang2022towards}, however, many of them are incompetent to catch up heuristic algorithms (such as the LKH for the traveling salesman problem) in terms of solving efficiency and decision quality. More importantly, they are not directly applicable to the optimization tasks under uncertainty, especially for unknown coefficients.
Therefore, a newer area of interest involves combining machine learning with optimization techniques to tackle problems with uncertain parameters, elaborated as follows.

\cite{bertsimas2020predictive} pioneered efforts in merging predictive and prescriptive analysis for optimization under uncertainty. A notable contribution is the SPO method~\cite{elmachtoub2022smart}, which introduces subgradient-based surrogate functions to replace non-differentiable regret functions in linear optimization problems, and its extension, SPO-relax~\cite{mandi2020smart}, which applies continuous relaxation to combinatorial problems and is utilized in our experiments. It is important to note that these studies primarily focus on predictive models before optimization solvers, which are often treated as standard heuristics (such as LKH3~\cite{helsgaun2017extension} or Dijkstra~\cite{dijkstra1959note}) or commercial solvers (like Gurobi~\cite{gurobi2019llc}).

Recently, there has been a growing body of work on predict-and-optimize approaches (also known as decision-focused learning~\cite{wilder2019melding,mandi2022decision,shah2022decisionfocused}). 
A significant trend within this category involves using relationships between solutions to optimization problems to develop better predictive models. For instance, the NCE (Noise-Contrastive Estimation) method~\cite{ijcai2021p0390} utilizes a noise-contrastive estimation technique~\cite{gutmann2010noise} to produce predictions that aim to achieve optimal solutions with better decision quality compared to non-optimal ones. The LTR (Learning to Rank) technique~\cite{mandi2022decision} explores the connection between pairwise learning to rank in NCE, leading to the development of various learning-to-rank methodologies such as pointwise rank~\citep{caruana1995using}, pairwise rank~\citep{joachims2002optimizing}, and listwise rank~\citep{cao2007learning}, which aim to generate predictions reflecting the relative importance of multiple solutions.

Another recent approach involves using neural network functions as surrogates for the original objective functions. Studies such as LODL~\cite{shah2022decisionfocused} and EGL~\cite{shah2024leaving} propose learning surrogate objective functions from sample data. LANCER~\cite{zharmagambetov2023landscape} adopts a similar strategy, integrating surrogate function learning with optimization solving and objective function learning. SurCO~\cite{ferber2023surco} suggests replacing the original non-linear objective with a linear surrogate, allowing the use of existing linear solvers.

In addition, some works propose implicit differentiation to obtain derivatives for convex optimization objectives for continuous optimization problems. OptNet~\cite{amos2017optnet} is the first work to differentiate quadratic programs, and then CPLayer~\cite{agrawal2019differentiable} and more~\cite{donti2017task,lee2019meta,eisenberger2022unified} extends this spirit to more general convex programs.
QPTL~\cite{wilder2019melding} proposes to differentiate through combinatorial optimization tasks by relaxing the problem to the continuous pace, and IntOpt~\cite{mandi2020interior} improves the interior-point method so that its backpropagation differentiable.
The following works ~\cite{wang2020automatically,ferber2020mipaal,paulus2021comboptnet} generate this paradigm for discrete problems.

However, some of these methods are limited to specific types of predict-and-optimize problems, such as quadratic optimization objectives~\cite{amos2017optnet} or convex objectives. Although methods based on solution importance~\cite{ijcai2021p0390,mandi2022decision} and surrogate objective functions~\cite{shah2022decisionfocused,shah2024leaving,zharmagambetov2023landscape} do not restrict the type of optimization, they require additional information, such as multiple solutions or a large number of optimization samples~\cite{shah2022decisionfocused}, to train the surrogate function before complete end-to-end learning can be performed. Furthermore, \textbf{the most critical issue} is that most of these methods struggle with combinatorial optimization due to the difficulty of differentiating through discrete decision variables, making predict-and-optimize for CO problems particularly challenging.

Besides the works mentioned above, some works also focus on the optimization under uncertainty. However, since current PnO methods do not support arbitrary uncertainty scenarios and are limited to cases with a known closed-form optimization objective but unknown coefficients, they are currently out of the scope of our benchmark. In ~\cite{joe2020deep,zhou2023reinforcement}, the form of the optimization objective function is uncertain and changes with time (especially for the pickup task), and \cite{tong2021usco} focuses more on the performance of regression among combinatorial spaces when there is no distribution of test data during training, but training data generation is available. Therefore, in this work, we mainly focus on methods within the predict-and-optimize paradigm and do not include them for now.

\section{Model Details}\label{suppl:model-detail}
\subsection{Prediction Model}
For a fair comparison, all end-to-end training models, as well as the two-stage approach, utilize the same multi-layer perception (MLP) for the prediction of optimization coefficients. The prediction model $\mathcal{M}$ using MLP is:
\begin{equation}
\mathbf{a}^{(i+1)} = \sigma(\mathbf{W}^{(i)} \mathbf{a}^{(i)} + \mathbf{b}^{(i)}
), \quad i = 1, 2, \ldots, K-1 ,
\end{equation}
where $\mathbf{a}^{(1)} = \mathbf{x} $ and $\mathbf{y} = \mathbf{a}^{(K)}$ are input and output for $\mathcal{M}$, $a^{i}$ is the hidden vector for $i=2,\cdots, K-1$, $\mathbf{W}$ is the weight term, $b$ is the bias term and $\sigma$ is the activation function where we adopt ReLU. In experiments, we adopt $K=3$ layers and the size of intermediate hidden units is set as 32.

\subsection{Two-stage approach in PtO} 
The two-stage approach is the vanilla model used when the coefficients of the optimization problem are uncertain and require prediction. In the prediction stage, the predictor estimates the coefficients of the optimization problem and optimizes with the original prediction loss using pre-collected labeled data pairs $D=\{x_i,y_i\}$. In the optimization stage, the coefficient on the test data is first inferred so that the optimization problem becomes certain. The vanilla solver is used to solve the optimization problem.

The final loss used in the two-stage approach is the same as the prediction stage, which includes the mean squared error (MSE) loss or binary cross entropy (BCE) loss of the prediction stage. 
\begin{equation}
    \begin{aligned}
        \operatorname{MSE}(\hat{y},y) &= \frac{1}{n} \sum_{i=1}^{n} (y_i - \hat{y}_i)^2,\\
        \operatorname{BCE}(\hat{y},y) &= -(y \log (a^{(K)})+(1-y) \log (1-a^{(K)})),
    \end{aligned}
\end{equation}
where $a^(K)$ is the output logits for the final layer neural network, $n$ is the number of training samples in MSE. The specific loss selection depends on whether the prediction task is regression (with MSE) or binary classification (with BCE). The used final loss is listed in Table~\ref{tab:data-all}.

\subsection{Details for PnO models}\label{suppl:pno-details}
\subsubsection{The Discrete category}
Table~\ref{tab:loss-discrete} lists the gradient interpolations for the discrete category.

\looseness=-1 A simple solution is to directly pass the gradient of the predicted coefficients from the objective, namely decision-focused learning (\textbf{DFL}) following~\cite{shah2022decisionfocused}. This spirit is also present in the straight-through estimator (\textbf{STE})~\cite{bengio2013estimating} to estimate the gradient of the ReLU activation function.
The representative work by gradient interpolation is \textbf{Blackbox}~\cite{poganvcic2019differentiation} which proposes linear interpolation by conducting a minor perturbation of original coefficient $\hat{\mathbf{y}}$ to produce informative gradients.
More recently proposed \textbf{Perturb}~\cite{berthet2020learning} and \textbf{I-MLE}~\cite{niepert2021implicit} adopt the same spirit to obtain informative gradients, and \textbf{Identity}~\cite{sahoo2022backpropagation} ignores the constraints and treats the gradient as a negative identity matrix.


\subsubsection{The Continuous category}
Popular approaches use the differentiation of Karush-Kuhn-Tucker (KKT) conditions proposed in \textbf{OptNet}~\cite{amos2017optnet} to obtain $\partial \mathbf{z}/\partial \mathbf{y}$ for the quadratic program. The later works extend this approach to cone programs~\cite{agrawal2019differentiating} (\textbf{cvxpy layer}) and more~\cite{donti2017task,lee2019meta, eisenberger2022unified}. However, these methods do not consider the case for discrete variables where gradients are blocked.
Several studies~\cite{wilder2019melding,mandi2020interior,wang2020automatically,ferber2020mipaal,paulus2021comboptnet} have extended the KKT differential approach to discrete problems. \textbf{QPTL}~\cite{wilder2019melding} extends this branch further to address singular gradient issues in discrete linear optimization by adding a squared norm regularizer.
\textbf{IntOpt}~\cite{mandi2020interior} addresses the challenge by differentiating homogeneous self-dual formulation for mixed integer programs to overcome difficulties of constraint satisfaction in computing the gradients using KKT conditions.

Alternatively, another branch of research~\cite{mandi2020smart,elmachtoub2020decision} has focused on adapting subgradient approximation methods from continuous linear problems to discrete scenarios.
\textbf{SPO-relax}~\cite{mandi2020smart} relaxes the problem and utilizes the surrogate \textbf{SPO+} loss proposed in ~\cite{elmachtoub2022smart}, and use the subgradient for backpropagation.
\begin{equation}
 L_{spo}(\mathbf{y}, \hat{\mathbf{y}})= -f(\mathbf{z}^\star(2 \hat{\mathbf{y}}-\mathbf{y}), 2 \hat{\mathbf{y}}-\mathbf{y} )+2 f(\mathbf{z}^\star(\mathbf{y}), \hat{\mathbf{y}})  - f(\mathbf{z}^\star(\mathbf{y}), \mathbf{y})~.
\end{equation}

The above categories propose on-the-fly gradient estimates to make gradients accessible. On the contrary, the subsequent two categories bypass the gradient approximation to train PnO.

\subsubsection{The Statistical category}
Table~\ref{tab:loss-statistical} lists the designed loss functions for the statistical category.

The representative method \textbf{NCE}~\cite{ijcai2021p0390} designs a noise-contrastive estimation (NCE)~\cite{gutmann2010noise} approach to produce predictions such that optimal solutions gain better decision quality than non-optimal ones.
\textbf{LTR}~\cite{mandi2022decision} discovers the inherent correlation between pairwise learning to rank and NCE, leading to the proposition of a series of learn-to-rank methodologies, including pointwise rank~\citep{caruana1995using}, pairwise rank~\citep{joachims2002optimizing}, and listwise rank~\citep{cao2007learning} loss to produce predictions that capture the relative importance of multiple solutions. 

\begin{table}[tb!]
\caption{Gradient interpolations for ``discrete" category where $\lambda$ is hyperparameter in Blackbox controlling trade-off between ``faithfulness'' and ``informativeness'', $\mathbf{I}$ is identity matrix, $\tau>0$ is temperature and $R$ is a sampled noise vector.}
\label{tab:loss-discrete}
\centering
\scalebox{.87}{
\begin{tabular}{ll}
\toprule
Method & Surrogate gradient \\\midrule
 Blackbox~\cite{poganvcic2019differentiation}  &   $\frac{\partial L}{\partial \mathbf{y}}=\frac{1}{\lambda}\left[\mathbf{z}(\hat{\mathbf{y}}+\lambda \frac{\partial L}{\partial \mathbf{z}}(\hat{\mathbf{z}}))- \mathbf{z}(\hat{\mathbf{y}})\right]$    \\
 Identity~\cite{sahoo2022backpropagation}  &   $\frac{\partial \mathbf{z}}{\partial \mathbf{y}}=-\mathbf{I} $   \\
 Perturb-and-MAP~\cite{berthet2020learning}      &   $\frac{\partial L}{\partial \mathbf{y}}= \mathbf{z}\left(\hat{\mathbf{y}}+\tau R\right)-\mathbf{z}(\mathbf{y}) $   \\
 I-MLE~\cite{niepert2021implicit} &   $\frac{\partial L}{\partial \mathbf{y}}=\mathbf{z}(\hat{\mathbf{y}}+\lambda \frac{\partial L}{\partial \mathbf{z}}(\hat{\mathbf{z}})+R)-\mathbf{z}(\hat{\mathbf{y}}+R) $   \\
\bottomrule
\end{tabular}%
}
\end{table}

\begin{table}[tb!]
\caption{Loss functions for ``statistical" category, where $p_\tau(\mathbf{z}\mid \mathbf{y})={\exp (-\frac{f(\mathbf{z}, \mathbf{y})}{\tau})}/{\Sigma_{\mathbf{z}^{\prime} \in S} \exp(\frac{-f\left(\mathbf{z}^{\prime}, \mathbf{y}\right)}{\tau})}$, $(p,q)$ is in ordered pairs $\operatorname{OP}$ means  $f(\mathbf{v}_p, \mathbf{y})<f(\mathbf{v}_q, \mathbf{y})$, the positive parameter $\nu$ is the margin, and $S$ represents the set of feasible solutions.}
\label{tab:loss-statistical}
\centering
\scalebox{0.9}{
\begin{tabular}{ll}
\toprule
Method & Designed loss function \\\midrule
 NCE~\cite{ijcai2021p0390}      &   $\Sigma_{\mathbf{z}_s \in S}(f(\mathbf{z}^\star, \hat{\mathbf{y}})-f(\mathbf{z}_s, \hat{\mathbf{y}})) $   \\
 pointwise-LTR~\cite{mandi2022decision}      &  $\frac{1}{|S|} \Sigma_{\mathbf{z}_s \in S}(f(\mathbf{z}_s, \hat{\mathbf{y}})-f(\mathbf{z}_s, \mathbf{y}))^2$    \\
 pairwise-LTR~\cite{mandi2022decision}      &  $\Sigma_{(p,q) \in OP}(\nu+f(\mathbf{z}_p, \hat{\mathbf{y}})-f(\mathbf{z}_q, \hat{\mathbf{y}}))$    \\
 listwise-LTR~\cite{mandi2022decision}       &  $\frac{1}{|S|} \Sigma_{\mathbf{z}_s \in S} p_\tau(\mathbf{z}_s \mid \mathbf{y})(\log p_\tau(\mathbf{z}_s \mid \mathbf{y})-\log P_\tau(\mathbf{z}_s \mid \hat{\mathbf{y}}))$  \\ \bottomrule 
\end{tabular}%
}
\end{table}

\subsubsection{The Surrogate Category}
Recent works \textbf{LODL}~\cite{shah2022decisionfocused} and \textbf{EGL}~\cite{shah2024leaving} propose learning the surrogate of objective functions from a set of samples.
\textbf{LANCER}~\cite{zharmagambetov2023landscape} learns surrogate functions in a similar way, while it integrates optimization solving and learning of objective functions. 
\textbf{SurCO}~\cite{ferber2023surco} proposes to replace the original non-linear objective with a linear surrogate, thus enabling the utilization of existing linear solvers. 

\begin{table*}[tb!]
\caption{Sensitivity analysis on capacity, variable size, and generalization on capacity on Knapsack (Gen) evaluated by regret(
Best three: \red{red}, \orange{orange}, \blue{blue}.
}
\label{tab:sensitivity}
\scalebox{.58}{%
\begin{tabular}{llccccccccccccccc}
\toprule
   & & {\textbf{Predictive}}  & & \multicolumn{3}{c}{\textbf{Discrete}}   & & \multicolumn{2}{c}{\textbf{Continuous}} &  & \multicolumn{4}{c}{\textbf{Statistical}} & & {\textbf{Surrogate}}  \\ 
      \cline{3-3} \cline{5-7} \cline{9-10} \cline{12-15} \cline{17-17}
      
    \multicolumn{2}{c}{\textbf{Setting}}   & {Two-stage} && {DFL} & {Blackbox} & {Identity} && {CPLayer}     & {SPO+} && {NCE} & {point-LTR} & {pair-LTR} & {list-LTR} && {LODL} \\ \midrule
\multirow{3}{*}{\makecell[l]{Sensitivity\\ on Capacity\\(Figure.~\ref{fig:sensi-cap})}} & Capacity 30 & 6.595 & & 11.744 & 24.274 & 31.874 & & 24.769 & \blue{6.223} & & 13.438 & 6.402 & 7.820 & \red{6.031} & & \orange{6.044}\\
 & Capacity 60 & 3.188 & & 3.355 & 21.244 & 22.824 & & 23.588 & \red{2.933} & & 10.459 & 5.480 & 4.488 & \orange{2.960} & & \blue{3.030} \\
 & Capacity 90 & \blue{1.566} & & 2.499 & 7.325 & 14.985 & & 7.511 & \red{1.345} & & 7.945 & 3.901 & 2.221 & \orange{1.417} & & {1.773} \\
\midrule

\multirow{4}{*}{\makecell[l]{Sensitivity\\ on variable size\\(Figure.~\ref{fig:sensi-size})}}& Size 40  & \orange{6.242} & & 11.480 & 29.934 & 34.519 & & 29.835 & \red{5.940} & & 11.178 & 11.209 & 8.193 & 6.506 & & \blue{6.408} \\
& Size 60 & \orange{5.941} & & 10.134 & 25.719 & 32.576 & & 26.475 & \red{5.675} & & 10.598 & 13.980 & 8.377 & 7.056 & & \blue{5.996}\\
& Size 80 & \orange{5.828} & & 12.241 & 27.149 & 36.901 & & 27.324 & \red{5.438} & & 10.927 & 14.080 & 7.546 & 6.328 & & 
\blue{6.159} \\
& Size 100  & \orange{6.298} & & 10.280 & 25.954 & 30.501 & & 26.454 & \red{6.055} & & 11.328 & 15.949 & 7.593 & 7.115 & & \blue{6.380} \\
\midrule

\multirow{3}{*}{\makecell[l]{Generalization\\on capacity\\(Figure.~\ref{fig:sensi-generalize})}} & {30-> 60} & 3.197 & & 4.848 & 23.913 & 25.046 & & 3.197 & 3.197 & & 12.312 & \red{2.995} & 5.331 & \blue{3.091} & & \orange{3.065}\\
 & {30->90} & \orange{1.716} & & 2.214 & 15.790 & 21.919 & & \orange{1.716} & 2.513 & & 25.970 & \red{1.694} & 14.010 & 3.047 & & \blue{1.723}\\
 &{30->120} & 0.644 & & \orange{0.303} & 8.724 & 30.709 & & 0.644 & 7.151 & & 34.304 & \blue{0.354} & 23.646 & 10.727 & & \red{0.152} \\
\bottomrule
\end{tabular}
}
\end{table*}

\section{Problem and Dataset Details}\label{suppl:data-detail}
For most problems, we adopt generated data or data that has been publicly available. For our new benchmark combinatorial advertising problem, which will be released, we collect data that contain encrypted and processed personal features and mobile application activity records, and all the used data are permitted by the users.
In the following, we elaborate on the detailed data processing procedure for each problem.

\subsection{Energy-cost aware Scheduling}\label{supp:prob-schedule}
With the more adoption of clean energy sources, energy demand and prices have become more adaptable~\cite{rolnick2022tackling}. 
In the context of industrial production, optimizing scheduling tasks based on energy prices has the potential to significantly conserve energy and reduce operational costs.

\textbf{Prediction}: 
The task is to predict the energy price of the 48 time slots (30 minutes each) given relevant features, including weather estimates, temperature, wind energy production, etc.

\looseness=-1 \textbf{Optimization}: The objective is to minimize the total energy cost of the schedule when the $J$ jobs are scheduled on $M$ machines and abide by the earliest start time and latest end time constraint.

{Suppose there are $M$ machines to deploy $J$ jobs with $R$ resources, and each day is split in $T$ equal timeslots ($T=48$ in our case).
Each job comes with its earliest start time $e_{j}$, latest end time $l_j$, duration $d_j$ and power usage $p_j$.
$u_{jr}$ is resource usage of task $j$ on resource $r$.
$c_{mr}$ denotes the capacity of machine $m$ for resource $r$.
The optimization objective is to minimize a linear program that completes scheduling tasks and reduces the energy cost:
\begin{align}
\min_{\mathbf{z}_{j m t}}~& \Sigma_{j \in J} \Sigma_{m \in M} \Sigma_{t \in T} \mathbf{z}^{j m t}\left(\Sigma_{t \leq t^{\prime}<t+d_j} p^j \mathbf{y}^{t^{\prime}}\right) \\
&\Sigma_{m \in M} \Sigma_{t \in T} \mathbf{z}^{j m t}=1, \forall_{j \in J} \label{eq:energy1}\\
&\mathbf{z}^{j m t}=0 \quad \forall_{j \in J} \forall_{m \in M} \forall_{t<e_j} \label{eq:energy2}\\
&\mathbf{z}^{j m t}=0 \quad \forall_{j \in J} \forall_{m \in M} \forall_{t+d_j>l_j}\label{eq:energy3}\\
&\Sigma_{j \in J} \Sigma_{t-d_j<t^{\prime} \leq t} \mathbf{z}^{j m t^{\prime}} u_{j r} \leq \mathbf{y}^{m r}, \forall_{m \in M} \forall_{r \in R} \forall_{t \in T}  \label{eq:energy4}
\end{align}
where $\mathbf{z}^{jmt}$ is a binary decision variable which is $1$ if task $j$ starts at time $t$ on machine $m$, else $0$.
Constraint~(\ref{eq:energy1}) ensures each task is scheduled once and only once. \textit{Constraint}~(\ref{eq:energy2}) and \textit{Constraint}~(\ref{eq:energy3}) make sure the job starts after the earliest start time and ends before the latest end time, and  \textit{Constraint}~(\ref{eq:energy4}) indicates the job does not exceed the capabilities of machines. {The data of each day is treated as one optimization instance.} We adopt $N=3$ machines, $R=1$ resources. The resource usage $u_{jr}$ is certain and given.}

\textbf{Dataset}: The data adopts open-sourced Irish Single Electricity Market Operator (SEMO)~\cite{ifrim2012properties} collected from Midnight 1st November 2011 to 31st December 2013. 

The price that needs to be predicted at each timeslot includes a 9-dimension vector as the input feature: calendar attributes, day-ahead estimates of weather characteristics, SEMO day-ahead forecasted energy load, wind-energy production and prices, and actual wind speed, temperature, CO$_2$ intensity, and price.

\textbf{License}: We adopt the publicly available dataset SEMO\footnote{https://www.sem-o.com/about/\label{foot:semo}}, which is licensed\footnote{https://www.sem-o.com/web-accessibility/} and regulated cooperatively by the Commission for Regulation of Utilities (CRU) in Ireland and the Utility Regulator for Northern Ireland (UREGNI, previously named NIAUR).

\subsection{Knapsack} 
The knapsack problem under uncertainty finds its applications, such as when an agent aims to efficiently carry items while reducing transportation costs and energy consumption, where the energy cost is unknown, which describes the following knapsack (energy) problem.

\textbf{Prediction}: 
The task is to predict the $j$-th item's value $\mathbf{y}^{j}$ from feature vector $\mathbf{x}^{j}$ for each $N$ item.

\textbf{Optimization}:
The optimization task is to maximize the total value of items in the knapsack with a capacity constraint, formulated as an integer linear objective:
\begin{equation}
\mathbf{z}^{\star}(y) = ~ \underset{\mathbf{z}}{\arg\max} \Sigma_{j=1}^N \mathbf{y}^j \mathbf{z}^j \quad
\text{s.t.} ~~ \Sigma_{j=1}^N \mathbf{w}^j \mathbf{z}^j \leqslant C ~.
\end{equation}
where $\mathbf{z}^j\in \{0,1\}$ is the binary variable indicating whether the item is picked, $\mathbf{w}^j$ is the weight for each item, and $C$ is the total capacity.

\textbf{Dataset}: 

\begin{itemize}
    \item \textbf{Synthetic} A synthetic dataset $\{\left(\mathbf{x}_1, \mathbf{y}_1\right),\left(\mathbf{x}_2, \mathbf{y}_2\right), \ldots,\left(\mathbf{x}_n, \mathbf{y}_n\right)\}$ is generated by the polynomial function following previous literature~\cite{elmachtoub2022smart}:
    \begin{equation}
    \mathbf{y}_i = [\frac{1}{3.5^{\operatorname{deg}} \sqrt{p}}\left(\left(\mathcal{B} \mathbf{x}_i\right)+3\right)^{\operatorname{deg}}+1 ] \cdot \epsilon_{i} ,
    \end{equation}
    where each $x_i \sim N(0, I_p)$ is generated from a multivariate Gaussian distribution, matrix $(\mathcal{B}^* \in \mathbb{R}^{d\times p} $ encodes the parameters of the true model, whereby each entry of $\mathcal{B}^\star$ is a Bernoulli random variable that is equal to 1 with probability 0.5, $\epsilon_{i}^j$ is a multiplicative noise term with uniform distribution, and $p$ is given number of features.
    The weights of the knapsack problem are regarded as certain and are sampled uniformly between 3 and 8.
    In our experiments, we set the default capacity as 30, and the number of items is 20. We use the polynomial degree $\operatorname{deg}$ of 4.
    
    \item \textbf{Real} The real data for knapsack adopts the SEMO dataset~\cite{ifrim2012properties}\textsuperscript{\ref{foot:semo}} in the previous energy-cost aware problem, where the energy price of each time slot is seen as profit, and the resource usage $u_{jr}$ of each time slot is seen as the weight of each item. which is certain and given.

\end{itemize}

\subsection{Budget Allocation}
The budget allocation problem under uncertainty has applications in scenarios where nonprofits aim to disseminate philanthropic information across multiple websites within the constraints of a total budget.

\textbf{Prediction}: 
Given the features $\mathbf{x}^w$ associated with the website $w$, the task is to predict $\mathbf{y}^{w}$, the probability that the information on the website $i$ will reach the customers. 

\textbf{Optimization}:
The objective is to maximize the expected number of users that are reached by the website at least once:
\begin{equation}
\mathbf{z}^\star(\mathbf{y}) =\underset{\mathbf{z}}{\arg\max} \frac{1}{N} \Sigma_{u=0}^N\left(1-\prod_{w=0}^M\left(1-\mathbf{z}^w \cdot \mathbf{y}^{wu}\right)\right) ~~
\text{s.t.} ~ \Sigma_{w=0}^M \mathbf{z}^w \leqslant B~.
\end{equation}
where $N$ is the number of users and $M$ is the number of websites. 

\textbf{Dataset}: The data comes from Yahoo! Webscope Dataset~\cite{yahoo07} with labels $y_i$ for each user $i$. We adopt the multi-linear relaxation following previous literature~\cite{wilder2019melding} where the features are generated by multiplying a random matrix $\mathbf{A}\in\mathbb{R}^{N\times N}$, i.e. $\mathbf{x}_i=\mathbf{A} \mathbf{y}_i$.
In our experiments, we set 5 websites, and 10 users with the default budget 1 following ~\cite{wilder2019melding,shah2022decisionfocused}.

\textbf{License}: We obtain this dataset from Yahoo's publicly available data~\footnote{https://webscope.sandbox.yahoo.com}, which is intended for non-commercial use by academics and other researchers. This data complies with Yahoo's data protection standards~\footnote{http://privacy.yahoo.com}, including stringent privacy controls. Usage of the data must adhere to the Data Sharing Agreement and terms of use~\footnote{https://legal.yahoo.com/us/en/yahoo/terms/otos/index.html} by Yahoo.

\subsection{Cubic Top-K}
The top-k problem adopted from ~\cite{shah2022decisionfocused} finds application in the field of Explainable XAI literature~\cite{hughes2018semi,futoma2020popcorn}. Here, the task revolves around the identification of the most salient features within the predictive model, specifically denoted as top-k.

\textbf{Prediction}: Given a dataset $\{\mathbf{x}_i, \mathbf{y}_i\}$, where $\mathbf{x}_i\sim U[0,1]$ is sampled from a uniform distribution, and $\mathbf{y}_i$ is generated according to cubic polynomial as follows:
\begin{equation}
    \mathbf{y}_i=10 \mathbf{x}_i^3-6.5 \mathbf{x}_i ~.
\end{equation}

\textbf{Optimization}:
It aims to select the $k$ largest elements from $\mathbf{y}$:
\begin{equation}
\mathbf{z}^\star(\mathbf{y})=\arg \operatorname{topk}(\mathbf{y})~. 
\end{equation}

\looseness=-1 \textbf{Dataset}: The item is randomly generated by uniform distribution on interval $[0,1)$. We set up 50 items with a budget of 5. This dataset is generated from scratch and does not involve datasets by other entities.

\subsection{Bipartite Matching}
Graph matching in social networks has diverse applications, helping people find jobs and friends online. However, the edge relationships between nodes are sometimes unknown, necessitating the prediction of the edge connections prior to conducting matching. 
We formulate this problem following previous literature~\cite{wilder2019melding,ferber2020mipaal,mandi2022decision}.

\textbf{Prediction}: 
\looseness=-1 Given the features $\mathbf{x}^i, \mathbf{x}^j$ of two nodes in each pair of nodes $(i, j)$, the objective is to predict whether there exists an edge between these two nodes ($\mathbf{y}_{ij}$ is 1 if the link exists, otherwise 0). 
\begin{equation}
    \mathbf{y}^{ij} = \mathcal{M}([ \mathbf{x}^i, \mathbf{x}^j ]) ~.
\end{equation}

\textbf{Optimization}:
The optimization of matching formulates as a linear objective under a permutation constraint:
\begin{equation}
\mathbf{z}^\star(\mathbf{y}) =\underset{\mathbf{z}}{\arg\max}~ \Sigma_{i=1}^N \Sigma_{j=1}^N \mathbf{y}^{ij} \mathbf{z}^{ij}  \quad
\text{s.t.}  \quad \mathbf{z} \bm{1}=\bm{1} \quad \mathbf{z}^{\top} \bm{1}=\bm{1} ~.
\end{equation}
where $\mathbf{y} \in \mathbb{R}^{N \times N}$ is the adjacency matrix and $\mathbf{z} \in \mathbb{R}^{N \times N}$ is the matching solution, and $\mathbf{z}^{ij} = 1 $ indicates that node $i$ matches $j$.

\textbf{Dataset}: We adopt the citation network Cora~\cite{mccallum2000automating,sen2008collective,yang2016revisiting} \footnote{http://www.cora.justresearch.com/} for the graph matching, where each node indicates a paper with a 1433-dimension feature obtained by bag-of-words methods.
We split the full graph into 27 instances following the previous literature first proposed in ~\cite{wilder2019melding} and later~\cite{ferber2020mipaal,mandi2022decision}, where each instance contains 100 nodes. Each instance constitutes a bipartite matching problem with cardinality 50.

\textbf{License}: The Cora dataset has been extensively utilized in numerous research studies in the domain of graph learning~\cite{kipf2016semi,velivckovic2017graph}. According to our findings~\footnote{https://www.openicpsr.org/openicpsr/project/100859/version/V1/view}, it is licensed under the Creative Commons Attribution 4.0 International (CC BY 4.0) License.

\subsection{Portfolio Optimization}
The allocation of assets plays a pivotal role in facilitating the circulation of funds across various areas of society and promoting economic efficiency.

\textbf{Prediction}: 
We use historical features such as daily price and volume data to predict the return $Y$ of the next day for $N$ stocks.

\textbf{Optimization}:
It aims to maximize the immediate return while reducing the risk with a classic Markowitz objective~\cite{markowitz2000mean}:
\begin{equation}
\mathbf{z}^\star(\mathbf{y})  =\underset{\mathbf{z}}{\arg\max}~~ \mathbf{z}^\top \mathbf{y}-\lambda \mathbf{z}^\top \mathbf{Q} \mathbf{z} \quad
\text{s.t.}  ~~ \Sigma_{i=0}^N \mathbf{z}^i = 1 ~.
\end{equation}
where $0 \leqslant \mathbf{z}^i \leqslant 1$ is a continuous variable indicating the fraction of money invested in security $i$, $\lambda=0.1$ is risk aversion constant, $\mathbf{Q} \in \mathbb{R}^{n\times n}$ is a positive semideﬁnite matrix representing the covariance between the returns of different securities. 

\looseness=-1 {\textbf{Dataset}: The data is obtained from public dataset S\&P500~\cite{quandl}, the collection of 505 largest companies in the US market from 2004 to 2017. The features include previous returns of 10 days, weeks, months, and years, as well as rolling averages of these time windows. 
We set the risk aversion parameter $\alpha=0.1$.}

\textbf{License}: We obtained the publicly available dataset from the website~\footnote{https://www.quandl.com/data/WIKI}, adhering to the following agreement~\footnote{https://data.nasdaq.com/terms} for its use.

\subsection{Combinatorial Advertising for Inclusive Finance}\label{suppl:data-adv-detail}
As previously mentioned, the new dataset on Combinatorial Advertising comprises real industry advertising records, wherein a fintech platform collaborates with financial institutions to offer low-interest loans to users. The advertisers promote their financial services through mobile applications (APP) utilizing a combination of various channels, including in-app notifications, text messages, telephone calls, and other methods.

\subsubsection{Predict-and-optimize formulation}
The prediction is elaborated in the main paper, and we add optimization form as follows:
\textbf{Optimization}: 
Denote $\mathbf{y}^{ij}$ the predicted conversion of user $i$ on strategy (advertising combination) $j$, the optimization objective is:
\begin{equation}
\begin{aligned}
 \mathbf{z}^\star(\mathbf{y}) =~&  \underset{\mathbf{z}}{\arg\max}~  \Sigma_{i=1}^N \Sigma_{j=1}^S \mathbf{y}^{ij} \mathbf{z}^{ij} \quad \\
 \text{s.t.}\quad~& \forall i, ~\Sigma_{j=1}^{S} \mathbf{z}^{ij} \leqslant 1 \quad\\
  &\Sigma_{j =1}^{S} \left({c}^j ~\Sigma_{i=1}^N \mathbf{z}^{ij} \right) \leqslant B ~.
\end{aligned}
\end{equation}
where $\mathbf{z}^{ij} \in\{0,1\}$ is 1 if marketing strategy $j$ is used for the user $i$. $S$ is the total set of strategies and $B$ is the total advertising budget. Note $|S| = 2^{C}$ where $C$ is the number of advertising channels (app message, text message, etc.) and strategy $j$ is the combination strategy of multiple channels. ${c}^j$ is the cost for each strategy.

\subsubsection{Data processing and model details}
The data processing can be encapsulated as follows:

\bb$\quad$ \textbf{Data Collection}: The data was collected by an industrial company in an advertisement for financial loans across 2023 in 203 days on 269329 users. 

\bb$\quad$ \textbf{Data Split}: The data has been split into distinct intervals, each spanning a period of 7 days. The dataset comprises a training set encompassing 161 days (comprising 23 instances) and a test set of 42 days (comprising 6 instances).

Within Each 7-Day Interval:
(a) User Filtering: A filtering process has been undertaken to select users who have engaged with at least two distinct marketing channels.
(b). Marketing Record Selection: For each marketing channel, the most recent marketing record is extracted and employed as input data.
(c) Feature Extraction: The features, including encrypted user feature vector and marketing records, encompassing hours 1 to 6, are transformed into feature vectors comprising 41 distinct features.

\bb$\quad$ \textbf{Prediction}: A four-layer neural network architecture has been employed as the predictive model, where the hidden units are 128, 64,32, and 1, respectively. The model's input comprises user features, historical marketing records (push), and marketing combinations, while the output is a prediction of user conversion rates within combination $j$.

\bb$\quad$ \textbf{Optimization}: For the Integer Linear Programming formulation,  we adopt OR-Tools solver~\cite{ortools} with SCIP (Strategic Consortium of Intelligence Professionals)~\cite{BestuzhevaEtal2021ZR} backend has been utilized for the resolution of the discrete linear programming problem.
Specific parameter values have been assigned to the optimization process, including channel costs (0.5 for channel 1, 1 for channel 2), costs for the combined strategy [0, 0.5, 1.0, 1.5] (which is simply treated as the sum of channel costs), and budget constraints, where the total budget is derived from the product of the number of users and the per-capital budget allocation of 0.1.

\subsubsection{Evaluation}\label{suppl:uplift}
As mentioned above, the coefficient $\mathbf{y}$ is not possibly available since we cannot observe the response to different treatments of the same user at the same time.
We refer to~\cite{goldenberg2020free,zhou2023direct}  and introduce 'uplift' as a metric for the improvement of advertising brought by the certain treatment, defined as the difference of the value of treatment group $P^\top$ with control group $P^C$:
\begin{equation}
\text{Uplift}=P^\top-P^C=P(\mathbf{y}=1 \mid \mathbf{w}=1)-P(\mathbf{y}=0 \mid \mathbf{w}=0)
\end{equation}
where $w$ indicates the treatment and $y$ indicates the label. In our case, the treatment group is the group of people whose combinatorial advertisement strategy is identical to the offline data, whereas the control group is the opposite.

\subsubsection{License, Use Terms and Privacy for combinatorial advertising dataset}\label{suppl:data-adv-lic}
If you use the combinatorial advertising dataset provided in the benchmark, you agree to the following terms of use from ``The Finvolution Group''.

\subsubsubsection{Data use terms and license}

\textbf{Acceptance of the Terms of Use}

The Finvolution Group strives to enhance public access to and use of data that it collects and publishes, to promote data utilization and academic research. The data are organized in datasets (the“Datasets”). The Datasets are collections of data, managed by The Finvolution Group and provided in a number of machine-readable formats. The Finvolution Group provides you with access to the Datasets free of charge subject to the terms of this agreement (these ``Dataset Terms''). Use of data derived from the Datasets, which may appear in formats such as tables and charts, is also subject to these Dataset Terms.



\subsubsubsection{Licenses and Restrictions}

You shall use the Datasets only for non-commercial research and educational purposes. You may provide research associates and colleagues with access to the Datasets provided that they first agree to be bound by these terms and conditions. If you are employed by a for-profit, commercial entity, your employer shall also be bound by these terms and conditions, and you hereby represents that he or she is fully authorized to enter into this agreement on behalf of such employer.

\textbf{No Association}

You may not use the name, any trade-mark, official mark, official emblem or logo of The Finvolution Group, or any of its other means of promotion or publicity, without The Finvolution Group's prior written consent nor in any event to represent or imply an association or affiliation with The Finvolution Group.

\textbf{No Warranties}

The Finvolution Group reserves the right at any time and from time to time to modify or discontinue, temporarily or permanently, this website, the Datasets, any means of accessing or utilizing the Datasets, or the API, at our sole discretion with or without prior notice to you.

The Finvolution Group may at our sole discretion, under any circumstances, for any or no reason whatsoever and with or without prior notice to you, terminate your access to the Datasets, any means of accessing or utilizing the Datasets or the API.

THE FINVOLUTION GROUP DISCLAIMS ALL WARRANTIES OF ANY KIND RELATED TO THE PROVISION OF THE DATASETS AND THE APIs.

\textbf{Exclusion of Liability}

THE FINVOLUTION GROUP SHALL NOT BE RESPONSIBLE OR LIABLE TO YOU FOR ANY LOSS OR DAMAGE OF ANY SORT INCURRED BY YOU IN CONNECTION WITH YOUR USE OF THE DATASETS. The Finvolution Group also shall not be responsible or liable for the accuracy, usefulness or availability of any data in the Datasets. 

You acknowledge that these Dataset Terms constitute a non-exclusive agreement. The Finvolution Group may develop products or services that compete with products or services that you offer without incurring any liability.

Other parties may have ownership interests in some of the data and information (“Materials”)  contained on the Site. The Finvolution Group in no way represents or warrants that it owns or controls all rights in all Materials, and The Finvolution Group will not be liable to you for any claims brought against you by third parties in connection with your use of any Materials.

These Dataset Terms may be amended by The Finvolution Group from time to at our sole
discretion. If there is a Chinese version of the Dataset Terms, the Chinese Dataset Terms shall prevail where the Chinese Dataset Terms conflict with the English Terms.

By continuing to use the Datasets subsequent to The Finvolution Group making available an amended version of these Dataset Terms, you acknowledge, agree and consent to such amendment.

No agency, partnership, joint venture, employee-employer, or franchiser-franchisee relationship is intended or created by these Terms and Conditions.

\subsubsubsection{Privacy}

Safeguarding user information and protecting user privacy is paramount in The Finvolution Group's operation since its inception. The Company has established a comprehensive administrative mechanism and standardized employee training system for stringent information security management. The Finvolution Group has also been deploying innovative technologies to promote user data protection. For example, the Company launched a Smart Finance Institute in 2018 for research and development in the field of artificial intelligence that can be applied in various aspects of financial services. In addition, The Finvolution Group is also a member of the National Information Security Standardization Technical Committee and Mobile Application (APP) Security Committee, maintaining up to date knowledge and compliant regarding the latest cyber-security regulatory requirements.

\section{Experimental Details}\label{suppl:exp-detiail}

\subsection{Analysis on Benchmark Results}\label{suppl:bench-res}
On the knapsack (gen) and knapsack (energy) problems, SPO+, pointwise and listwise LTR, and LODL achieve lower regret than the two-stage method.
For the energy scheduling problem, SPO+, LODL, and all statistical methods except pointwise-LTR exceed the two-stage approach.
In the budget allocation problem, the LODL surprisingly does not outperform the two-stage approach, probably because it could be hard to learn a surrogate function for the non-linear objective.
In the cubic top-K problem, optimal performance is achieved by the two-stage approach, which differs from the results of previous works~\cite{shah2022decisionfocused} since a two-layer MLP is used as the predictive model. This may be attributed to the inherent simplicity of the sampled prediction data. 
The bipartite matching problem exhibits high regret on all models, probably due to the lack of samples in the train and test set, such that the over-smoothing issue may cause performance degradation.
Most models have high regret for portfolio optimization, whereas pointwise-LTR and LODL have lower regret levels than the two-stage approach.
In the combinatorial advertising problem, we do not run statistical class methods since the decision variables differ for processed instances, making the solution cache not applicable. For the surrogate class, the samples required to learn the surrogate objective function are hard to obtain. The suboptimal outcomes observed in continuous class methods may stem from the intricate nature of user behavior, posing challenges for the gradient backpropagation of convex functions.

\begin{figure*}[tb!]
    \centering
    \begin{tabular}{ccc}
        \includegraphics[width=0.28\textwidth]{figs/size/knapsack-gen_new-size40.pdf}
        &\includegraphics[width=0.28\textwidth]{figs/size/knapsack-gen_new-size60.pdf} 
        &\includegraphics[width=0.28\textwidth]{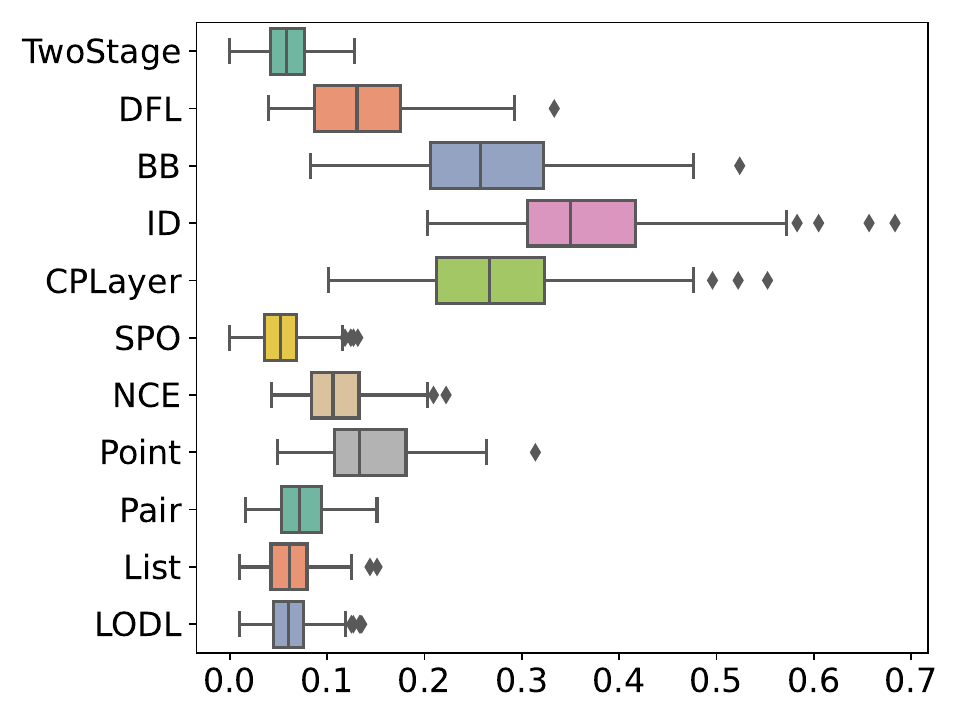}
        \\  
    (a) Variable size of 40 & (b) Variable size of 60  & (c) Variable size of 80
    \end{tabular}
    \caption{Regret on knapsack (gen) problem with increasing decision variable size (40, 60, 80). With an increasing decision variable number, performance fluctuates and degrades in some models (BB, ID, and pointwise-LTR).}
    \label{fig:sensi-size}
\end{figure*}

\begin{figure*}[tb!]
    \centering
    \begin{tabular}{ccc}
        \includegraphics[width=0.28\textwidth]{figs/gen/knapsack-gen_gen60.pdf}
        &\includegraphics[width=0.28\textwidth]{figs/gen/knapsack-gen_gen90.pdf} 
        &\includegraphics[width=0.28\textwidth]{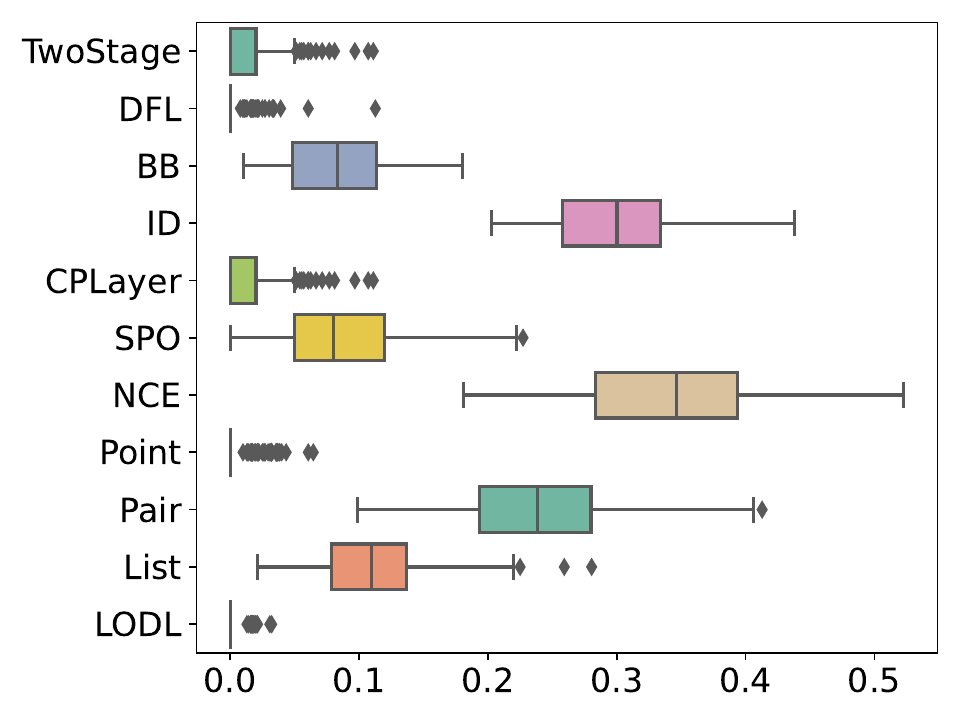}
        \\  
    (a) Generalize to capacity 60 & (b) Generalize to capacity 90  & (c) Generalize to capacity 120 
    \end{tabular}
    \caption{Regret on knapsack (gen) problem when generalizing trained model on capacity 30 to 60, 90 and 120. Most end-to-end trained models degrade in the generalization task.}
    \label{fig:sensi-generalize}
\end{figure*}

\begin{table*}[tb!]
\caption{ {Average train/test time per epoch (in second) with variable size 40, 60, 80, 100, 200 on Knapsack (Gen) Dataset. Note that LODL data generation time is not listed but gradually escalates, becoming the primary bottleneck.}
}
\label{tab:sensi-size-time}
\scalebox{.60}{%
\begin{tabular}{llccccccccccccccc}
\toprule
& & {\textbf{Predictive}}  & & \multicolumn{3}{c}{\textbf{Discrete}}   & & \multicolumn{2}{c}{\textbf{Continuous}} &  & \multicolumn{4}{c}{\textbf{Statistical}} & & {\textbf{Surrogate}}  \\ 

\cline{3-3} \cline{5-7} \cline{9-10} \cline{12-15} \cline{17-17}
      
\multicolumn{2}{c}{\textbf{Setting}}   & {Two-stage} && {DFL} & {Blackbox} & {Identity} && {CPLayer}     & {SPO+} && {NCE} & {point-LTR} & {pair-LTR} & {list-LTR} && {LODL} \\ \midrule

\multirow{2}{*}{\makecell[c]{Size 40}}  & Train Time &0.061 & & 1.634 & 1.628 & 2.121 & & 0.671 & 1.376 & & 2.516 & 5.566 & 17.954 & 4.490 & & 4.818  \\
 & Test Time &1.725 & & 2.141 & 2.888 & 7.078 & & 2.317 & 3.450 & & 4.228 & 5.404 & 30.909 & 8.122 & & 1.619  \\
\midrule

\multirow{2}{*}{\makecell[c]{Size 60}}  & Train Time  &0.078 & & 2.397 & 1.737 & 12.709 & & 0.575 & 1.585 & & 3.328 & 10.530 & 24.310 & 4.888 & & 6.562  \\
& Test Time &2.111 & & 2.878 & 2.439 & 44.366 & & 4.192 & 4.188 & & 8.221 & 10.787 & 51.412 & 11.987 & & 2.063  \\
\midrule

\multirow{2}{*}{\makecell[c]{Size 80}}  & Train Time  &0.882 & & 5.839 & 2.299 & 27.327 & & 0.649 & 3.170 & & 4.455 & 19.337 & 116.366 & 8.780 & & 9.190  \\
& Test Time &3.594 & & 4.703 & 3.969 & 108.740 & & 4.676 & 7.577 & & 16.063 & 18.999 & 69.713 & 21.053 & & 2.295  \\
\midrule

\multirow{2}{*}{\makecell[c]{Size 100}}  &  Train Time &1.025 & & 3.197 & 5.637 & 2.884 & & 0.636 & 4.158 & & 4.536 & 21.758 & 114.583 & 25.009 & & 11.318  \\
& Test Time &4.355 & & 3.968 & 8.579 & 6.486 & & 4.617 & 10.474 & & 14.705 & 20.370 & 493.726 & 47.784 & & 2.816  \\
\bottomrule
\end{tabular}
}
\end{table*}

\subsection{Details of versatility of PnO in RQ3}\label{supp:fails}
This section explores sensitivity on the constraint (i.e. 
 the capacity in knapsack problem) and variable size on the knapsack (gen) problem. We also show generalization results when the model is trained on the optimization problem of capacity 30 and tested directly on capacities 60 and 90. The detailed results for sensitivity analysis Table~\ref{tab:sensitivity}.

In Figure~\ref{fig:sensi-size}, we evaluate the knapsack problem by gradually increasing the number of variables (with constraints proportionally scaling accordingly) to observe changes in the model. The running time is also reported in Table~\ref{tab:sensi-size-time}. We set the decision variable size of 40, 60, and 80 and observe that certain models, such as BB, ID, and point-wise LTR, exhibit fluctuations and degradations to different extents. Some models demonstrated more stable performance, including the two-stage and statistical and surrogate approaches, which could be attributed to the reliance on the relative importance of multiple solutions, which is less sensitive to the change of decision variable numbers.

Lastly, in Figure~\ref{fig:sensi-generalize}, we conduct experiments to investigate the generalization performance by deploying models trained on the dataset of capacity 30  directly to problems with constraints of 60, 90, and 120. The figure indicates that certain models, including the two-stage model, LODL, and pointwise LTR, demonstrated the ability to prevent performance deterioration. Some models even exhibited improvements, such as the continuous-class models cvxpy layer and SPO. Conversely, several models experienced a decline in performance, notably the discrete-class Blackbox and Identity models, as well as statistical-class models like NCE and LTR.

\begin{table*}[tb!]
\caption{Sensitivity analysis on hyper-parameters (hidden units, number of training epochs) on Knapsack (Energy) evaluated by regret(
Best three: \red{red}, \orange{orange}, \blue{blue}.
}
\label{tab:sensi-hyper}
\scalebox{.64}{%
\begin{tabular}{llccccccccccccccc}
\toprule
   & & {\textbf{Predictive}}  & & \multicolumn{3}{c}{\textbf{Discrete}}   & & \multicolumn{2}{c}{\textbf{Continuous}} &  & \multicolumn{4}{c}{\textbf{Statistical}} & & {\textbf{Surrogate}}  \\ 
      \cline{3-3} \cline{5-7} \cline{9-10} \cline{12-15} \cline{17-17}
      
    \multicolumn{2}{c}{\textbf{Setting}}   & {Two-stage} && {DFL} & {Blackbox} & {Identity} && {CPLayer}     & {SPO+} && {NCE} & {point-LTR} & {pair-LTR} & {list-LTR} && {LODL} \\ \midrule
\multirow{3}{*}{\makecell[c]{Hidden\\units}}
& 8 & 9.456 & & 9.467 & 20.218 & 21.352 & & 20.430 & \red{7.824} & & 11.778 & 8.606 & \blue{8.534} & \orange{7.904} & & {9.339}\\

& 32 & 8.745 & & \blue{8.353} & 35.705 & 17.156 & & 36.402 & 8.407 & & 11.932 &
\orange{8.236} & 9.022 & \red{8.083} & & {9.567}\\

& 128 & 8.319 & & \blue{8.318} & 45.809 & 45.149 & & 45.948 & 8.979 & & 12.028 & \orange{8.317} & 8.758 & \red{7.794} & & {9.127}\\ \midrule

\multirow{3}{*}{\makecell[c]{Train\\epochs}}
&100 & 9.178 & & 9.108 & 35.705 & 22.013 & & 36.402 & \blue{8.407} & & 11.932 & \orange{8.236} & 9.840 & \red{8.083} & & {9.567} \\

& 300 & 8.745 & & \blue{8.353} & 35.705 & 17.156 & & 36.402 & 8.407 & & 11.932 & \orange{8.236} & 9.022 & \red{8.083} & & {9.567}\\

&500 & 8.745 & & \blue{8.353} & 35.705 & 16.826 & & 36.402 & 8.407 & & 11.932 & \orange{8.236} & 9.840 & \red{8.083} & & {9.567} \\

\bottomrule
\end{tabular}
}
\end{table*}

\begin{table*}[tb!]
\caption{Sensitivity analysis on the number of training samples) on Knapsack (Gen) evaluated by regret(
Best three: \red{red}, \orange{orange}, \blue{blue}.
}
\label{tab:sensi-hyper-samples}
\scalebox{.63}{%
\begin{tabular}{llccccccccccccccc}
\toprule
   & & {\textbf{Predictive}}  & & \multicolumn{3}{c}{\textbf{Discrete}}   & & \multicolumn{2}{c}{\textbf{Continuous}} &  & \multicolumn{4}{c}{\textbf{Statistical}} & & {\textbf{Surrogate}}  \\ 
      \cline{3-3} \cline{5-7} \cline{9-10} \cline{12-15} \cline{17-17}
      
    \multicolumn{2}{c}{\textbf{Setting}}   & {Two-stage} && {DFL} & {Blackbox} & {Identity} && {CPLayer}     & {SPO+} && {NCE} & {point-LTR} & {pair-LTR} & {list-LTR} && {LODL} \\ \midrule
\multirow{2}{*}{\makecell[c]{Num of\\samples}}& 400 & 6.595 & & 11.744 &  24.274 & 31.874 & & 24.769 & \blue{6.223} & & 13.438 & 6.402 & 7.820 & \red{6.031} & & \orange{6.044} \\

& 2000 & \blue{6.389} & & 25.017 & 6.595 & 6.595 & & 24.480 & \orange{6.168} & & 19.207 & \red{5.590} & 8.385 & 6.471 & & -\\

\bottomrule
\end{tabular}
}
\end{table*}

\subsection{Sensitivity analysis of training on hyperparameters}\label{suppl:exp-sensi}
This section explores the sensitivity of each model on hyper-parameters during training, as shown in Table~\ref{tab:sensi-hyper}.
To begin with, we explore the influence of hidden size on the ultimate decision. We observe that with increasing hidden size, the two-stage method's result improves slightly. Listwise-LTR remains a good performance at different hidden sizes, but SPO+ achieved the best score at the unit of 8, maybe because a larger neural network would cause SPO+ to overfit.
Then, we show the sensitivity results with respect to the number of training epochs. We observe that the two-stage method improves slightly from 100 to 300, while most methods have converged at the epoch of 300.
Last, we explore the sensitivity analysis with respect to the number of samples (400 and 2000 in the training set) on the knapsack (gen) dataset, shown in Table~\ref{tab:sensi-hyper-samples}. We observe that the two-stage method improves slightly, and SPO+ achieves the best results with more training samples. The performance of some PnO methods degrades slightly, probably due to instability by more samples. Note the result of LODL is not applicable since the data generation time exceeds the threshold.

\begin{table*}[tb!]
\caption{{Theoretical assumptions and insights of PnO in literature. ``LR", ``RF" and ``KNN" is short for linear regression, random forest, and K-nearest-neighbor.}}
\label{tab:cmp-theory}
\scalebox{0.81}{
\begin{tabular}{ccccl}
\toprule
Paper & \makecell[c]{Prediction\\ model}             & \makecell[c]{Optimization\\ form} & \makecell[c]{Perfect prediction\\ assumption} & Theoretical insight \\
\midrule
\cite{balkanski2017minimizing} & Not Specified & Submodular        & Required                      & \makecell[l]{Exists submodular functions that no algorithm can \\obtain an approximationstrictly better than\\$1/2 - O(1)$ using polynomially-many samples.} \\ \hline

\cite{chen2020optimization} & Not specified                & Submodular        & Required       
 &  \makecell[l]{With structure information, achieve a constant\\approximation for the maximum coverage problem } \\\hline

\cite{bertsimas2020predictive} & \makecell[c]{LR, RF, KNN, etc.} & All               & Required                      &  \makecell[l]{Under appropriate conditions and for certain\\predictive prescriptions, prescription converge to\\true full information optimizers } \\\hline

\cite{cameron2022perils} & All  & All       & Not required                  &  \makecell[l]{PnO is better than PtO by making the right error\\ trade-offs}   \\                                                       \bottomrule                                
\end{tabular}
}
\end{table*}

\subsection{Theoretical Interpretation on benchmarks}\label{suppl:theory}
Though this work is focused on benchmarking PnO problems, we also attempt to discuss the literature on PnO from a theoretical point of view. 
As shown in Table~\ref{tab:cmp-theory}, previous literature ~\cite{balkanski2017minimizing} discusses the hardness of solving submodular function when the optimization function is unknown, and \cite{chen2020optimization} proposes that under certain conditions (where structure information is given), a constant-approximation algorithm could be designed. 
In the context of bipartite matching problems within our framework, a correlation exists wherein the prediction of graph structure precedes the determination of matching. This could be helpful to this benchmark given the observed high decision regret.

Then, the recent work~\cite{bertsimas2020predictive} also discusses the capability of PtO. However, it is built on a strong assumption that the prediction is perfect without error, which makes it hard to be applicable to real scenarios. In the real world and in this work, we are more focused on the ``perfect prediction" of some distribution on the dataset and investigate how to propose and deploy end-to-end models that achieve better ultimate decisions when the prediction is not perfect.
\cite{cameron2022perils} proposes an insightful perspective that end-to-end training of PnO is better than the two-stage training of PtO even when the prediction is perfect, since PnO models make the right error trade-off. This perspective could also be validated partially by the experiment shown in Figure~\ref{fig:finetune}(c~d), where the prediction curve and decision curve perform differently by some models (e.g., Blackbox).

\subsection{Qualitative Analysis}\label{suppl:quali}

In Fig~\ref{fig:finetune}, we visually scale the prediction loss of Blackbox to $5*10^{9}$ for clarity, while others to $10^{8}$. 

\begin{figure*}[tb!]
    \centering
    \begin{tabular}{cc}
        \includegraphics[width=0.4\textwidth]{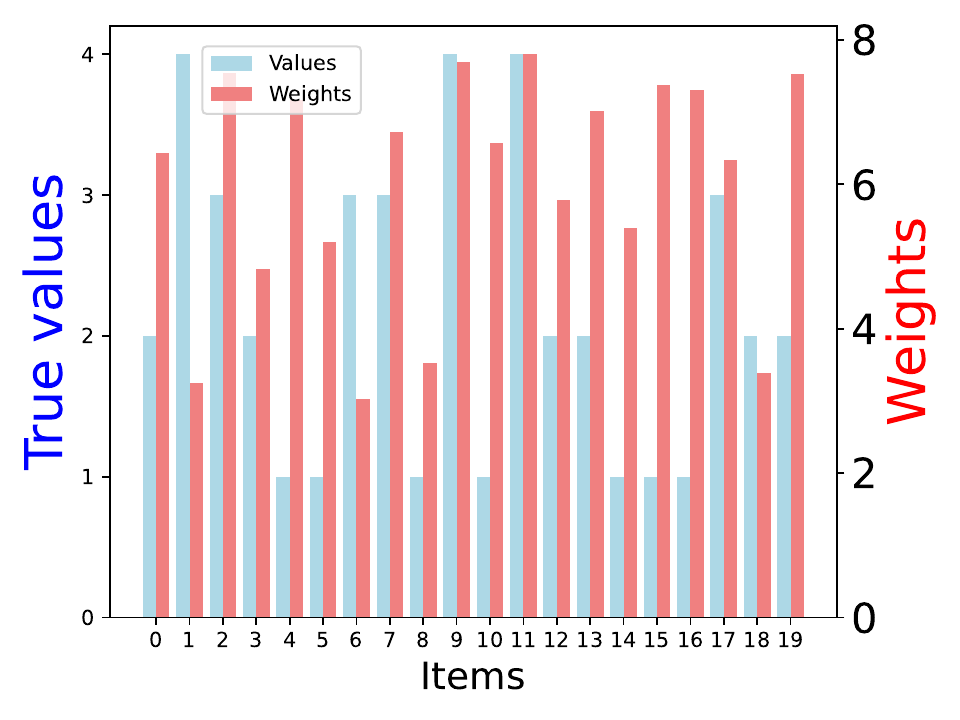}
        &\includegraphics[width=0.4\textwidth]{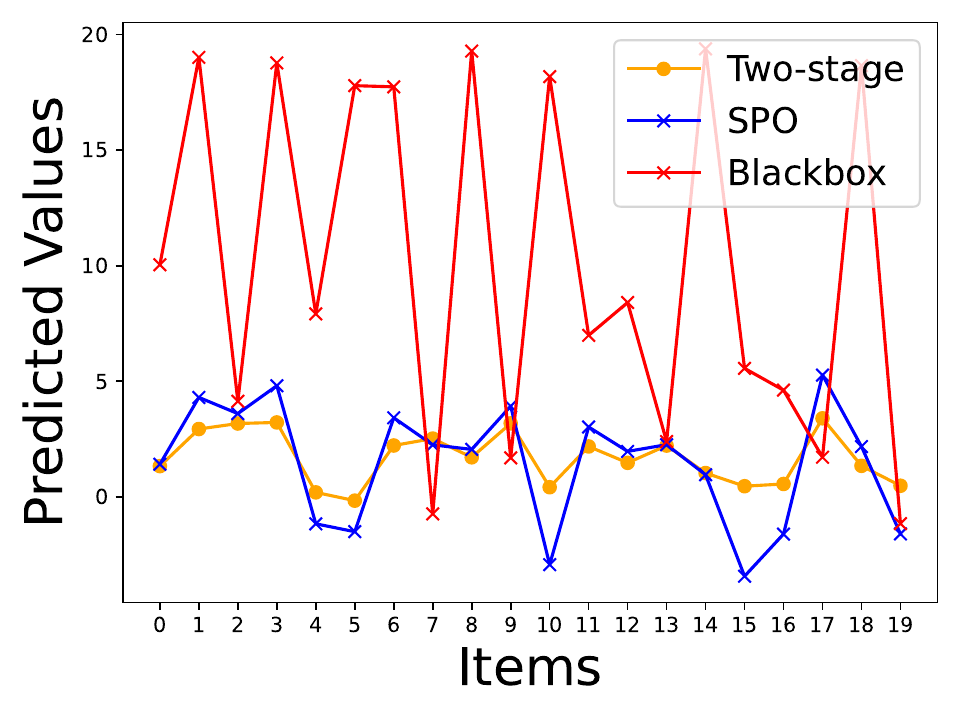} \\
    (a) Values and weights of knapsack problem & (b) Predicted values for various methods 
    \end{tabular}

    \caption{Qualitative analysis for knapsack problem on the 10-th test instance.}
    \label{fig:quali-kp}
\end{figure*}

We conduct a set of qualitative analyses on the knapsack problem to explore the reasons behind the success or failure of these methods. First, Fig~\ref{fig:quali-kp}(a) visualizes a bar plot of the 10-th test instance in the knapsack (gen) problem instance, showing the values and weights.  We can see that the range of predicted values by SPO is roughly the same as that of the two-stage model, but the Blackbox model, without labels, has a significant difference from the true values. In Fig~\ref{fig:quali-kp} (b) we plot the predicted values for 20 items for three models: the two-stage model, the well-performing SPO model, and the poorly-performing Blackbox model.

We can observe that the SPO model achieved the best decision quality (with the objective value of 18, selecting items 1, 3, 6, 9, 17, and 18). The two-stage model's decision quality was slightly lower (with the objective value of 17, selecting items 1, 3, 6, 8, 9, and 17). SPO improved decision quality by replacing item 9 with item 18, by predicting a higher predicted value to item 18, and by lowering the prediction for item 9, which is guided by information of optimization objective while the two-stage does not involve this. Although this increased the prediction error, it enhanced the decision quality.

For the Blackbox model, we observed that its predicted values were similar to SPO for some items, but due to the lack of label supervision during the prediction phase (not using coefficient labels), the prediction errors were larger, leading to poorer decision quality.

\section{Terms of Use, Limitations and Broader Impacts}\label{suppl:limits}

\textbf{Terms of Use} 
As part of our benchmark work, we utilized numerous datasets in the benchmarking process, and we adhere to the respective usage terms and licenses elaborated in Appendix~\ref{suppl:data-detail}.
Additionally, we are releasing a new industrial dataset, and the license, use terms, and privacy are also elaborated in Appendix~\ref{suppl:data-adv-lic}.
If readers use our benchmark dataset, please also adhere to the corresponding user terms specified by them.

\textbf{Limitations} 
The limitations of this paper include the inherent difficulty in analyzing the internal mechanisms of PnO. 
Aside from a few existing works that examine the theoretical underpinnings of PnO (like SPO~\cite{elmachtoub2022smart}), at present, it is challenging to provide theoretical explanations or straightforward criteria to determine when PnO outperforms PtO or which specific model is the best choice. Therefore, our focus for now is primarily on the empirical results of these methods on current datasets. To assist practitioners, we have categorized and analyzed these methods based on their generalizability, usage conditions, and associated costs, and our experimental results. 
In Appendix~\ref{suppl:theory}, we also attempt to interpret the benchmark results based on existing theories for the reader's reference.We offer a predict-and-optimize model using deployment recommendations in Sec.~\ref{sec:guide} based on these above considerations for practitioners.

Our work does not aim to cover all combinatorial optimization problems, predictors, solvers, or Predict-and-Optimize (PnO) methods. Instead, we present a user-friendly framework with representative implementations for each module. Readers who wish to implement their deployments in any module can do so by replacing the corresponding block within our framework.

\textbf{Broader Impacts} 
As far as we are concerned, this paper does not have critical negative societal impacts. We hope that this benchmark work can support better decision-making under uncertainty by utilizing end-to-end prediction and optimization methods for training predictive models more appropriately. 
In Appendix~\ref{suppl:data-detail}, we also discuss the possible positive implications of these decisions, including more energy-efficient production scheduling models, more rational resource allocation under uncertain scenarios, more appropriate social matching, more interpretable models, more appropriate capital allocation, and more inclusive financial services.
Certainly, we acknowledge that similar to previous works on predict-and-optimize (PnO), decision-making under uncertainty could potentially be applied to inappropriate purposes. However, given the current challenges of deploying PnO in industry and practice, we consider the negative impacts will be minimal. We hope that this benchmark can facilitate future advances in this field and aid in creating better decisions under uncertain scenarios for people.
